\documentclass{article}

\usepackage{nips14submit_09,times}

\usepackage[utf8]{inputenc} 
\usepackage[T1]{fontenc}    
\usepackage{url}            
\usepackage{booktabs}       
\usepackage{amsmath,amssymb}
\usepackage{amsfonts}       
\usepackage{nicefrac}       
\usepackage{microtype}      
\usepackage{algorithm}
\usepackage{algorithmic}
\usepackage{graphicx}
\usepackage{subcaption}

\newtheorem{proposition}{Proposition}
\newtheorem{lemma}{Lemma}

\DeclareMathOperator*{\E}{\mathbb{E}}
\DeclareMathOperator*{\V}{\mathbb{V}}
\DeclareMathOperator*{\argmax}{arg\,max}

\newcommand*{\QEDB}{\hfill\ensuremath{\square}}

\title{Merging Deterministic Policy Gradient Estimations with Varied Bias-Variance Tradeoff for Effective Deep Reinforcement Learning}

\author{%
  Gang Chen\\
  School of Engineering and Computer Science\\
  Victoria University of Wellington\\
  New Zealand\\
  \texttt{aaron.chen@ecs.vuw.ac.nz} \\
}

\begin{document}

\maketitle

\begin{abstract}
\emph{Deep reinforcement learning} (DRL) on \emph{Markov decision processes} (MDPs) with continuous action spaces is often approached by directly training parametric policies along the direction of estimated \emph{policy gradients} (PGs). Previous research revealed that the performance of these PG algorithms depends heavily on the bias-variance tradeoffs involved in estimating and using PGs. A notable approach towards balancing this tradeoff is to merge both on-policy and off-policy gradient estimations. However existing PG merging methods can be sample inefficient and are not suitable to train deterministic policies directly. To address these issues, this paper introduces \emph{elite PGs} and strengthens their variance reduction effect by adopting \emph{elitism} and \emph{policy consolidation techniques} to regularize policy training based on policy behavioral knowledge extracted from elite trajectories.  Meanwhile, we propose a two-step method to merge elite PGs and conventional PGs as a new extension of the conventional interpolation merging method. At both the theoretical and experimental levels, we show that both two-step merging and interpolation merging can induce varied bias-variance tradeoffs during policy training. They enable us to effectively use elite PGs and mitigate their performance impact on trained policies. Our experiments also show that two-step merging can outperform interpolation merging and several state-of-the-art algorithms on six benchmark control tasks.
\end{abstract}

\section{Introduction}
\label{sec-in}

Research on \emph{deep reinforcement learning} (DRL) algorithms has made remarkable progress in solving many challenging \emph{Markov Decision Processes} (MDPs) with continuous state spaces and continuous action spaces \cite{bhatnagar2009automatica,lillicrap2015,munos2016nips,mnih2016icml,gu2016icra,wang2016arxiv,wu2017nips}. Among all the state-of-the-art DRL algorithms, an important family aims at directly learning \emph{action-selection policies} modelled as \emph{deep neural networks} (DNNs) by iteratively applying \emph{policy gradient} (PG) updates to the DNNs based on either (or both) on-policy or (or and) off-policy samples collected from the learning environment. In comparison to DRL algorithms that indirectly derive policies from learned \emph{value functions}, PG algorithms have well-understood convergence properties and are naturally suited to handle continuous action spaces while achieving high sample efficiency on large-scale MDPs \cite{deisenroth2013}.

Previous research revealed that PG algorithms can exhibit highly unstable learning behavior due to to several major issues \cite{geist2019,ahmed2019,haarnoja2017,haarnoja2018icml,donoghue2017,osband20161,nachum2017,houthooft2018,kaplanis2019}. In particular, PG algorithms can be seriously affected by the bias-variance tradeoffs involved in estimating and using PGs \cite{tucker2018,schulman2015,papini2018,shi2019}. A notable approach towards tackling this problem is to merge multiple gradient estimations, as demonstrated in \cite{gu2017,gu2016,wang2016arxiv}. In fact researchers have explored the benefits of merging on-policy estimation of PG with off-policy estimation to unleash the advantages of both \cite{gu2017}. However, existing PG merging techniques can only train stochastic policies and may also suffer from poor data efficiency due to one-time use of environment samples collected in the on-policy fashion. They are hence unsuitable for pure off-policy PG algorithms such as \emph{Deep Deterministic Policy Gradient} (DDPG) \cite{lillicrap2015} and twin-delayed DDPG (TD3) \cite{fujimoto2018}, which are designed to train deterministic policies. In view of this, it is important to develop new algorithms to merge multiple off-policy estimations of \emph{deterministic PGs} (DPG) (i.e., PGs of deterministic policies) for enhanced learning efficiency and effectiveness. This research question has not attracted sufficient attention in the past.

We seek to investigate two inter-related questions: (Q1) how to estimate DPG in an off-policy fashion with varied bias-variance tradeoffs; and (Q2) how to merge multiple DPG estimations to effectively train a policy network. To answer Q1, we propose to estimate \emph{conventional DPG} and \emph{elite DPG} respectively based on two separate \emph{experience replay buffers} (ERBs), i.e., the \emph{full ERB} and the \emph{elite ERB}. The full ERB keeps track of all state samples obtained since the beginning of the learning process. The elite ERB maintains an elite group of sampled \emph{trajectories} with the highest \emph{cumulative rewards}. Hence, conventional DPG is expected to have lower bias but higher variance than elite DPG. On the other hand, the newly introduced elite DPG aims at exploiting and improving best trajectories sampled so far. To further reduce the estimation variance associated with elite DPG, inspired by \cite{kaplanis2019}, we develop a policy consolidation technique to extract policy behavioral knowledge from the elite ERB with the help of a generative model such as the \emph{variational autoencoder} (VAE) \cite{doersch2016}. Elite DPG is then constrained by the behavioral deviation of trained policies from the VAE model.

To answer Q2, we study two different DPG merging methods. The first \emph{interpolation merging} method linearly combines conventional DPG with elite DPG \cite{gu2017,gu2016}. The second \emph{two-step merging} method is newly developed in this paper to amplify the variance reduction effect of elite DPG through a two-step iterative process. Theoretically, we show that both interpolation merging and two-step merging enable us to reduce the variance of trained \emph{policy parameters} with controllable bias, contributing to improved learning reliability and performance. Meanwhile, the performance impact of using biased estimation of elite DPG can be effectively bounded. On six difficult benchmark control tasks, we show that TD3 enhanced by two-step merging (TD3-2M) can noticeably outperform TD3 enhanced by interpolation merging (TD3-IM), TD3 and other cutting-edge DRL algorithms, including Soft Actor-Critic (SAC) \cite{haarnoja2018icml}, Proximal Policy Optimization (PPO) \cite{schulman20171} and Interpolated Policy Gradient (IPG), which is designed to merge on-policy and off-policy PG estimations \cite{gu2017}.

\section{Related Work}
\label{sec-rw}

Merging multiple DRL methods has been frequently attempted in the literature. Q-Prop \cite{gu2016}, PGQ \cite{donoghue2017}, ACER \cite{wang2016arxiv} and IPG \cite{gu2017} are some of the recent examples with the aim to combine on-policy and off-policy learning.  Different from these research works, this paper studies the task of learning deterministic policies by merging multiple off-policy DPGs without involving on-policy learning. Our proposed use of the elite ERB is related to the prioritized experience replay (PER) technique \cite{horgan2018}. However, this paper aims to balance the bias-variance tradeoff via DPG merging rather than PER since PER introduces additional bias to all estimated DPGs, potentially affecting learning stability. Hence our research is more closely related to IPG than PER.

Huge efforts have been made to control noise in estimated PGs. For example, stochastic policies can be trained with reduced variance by using smoothed and target value function networks \cite{nachum2018,anschel2017,van2016}. In this paper, we rely on time-tested variance reduction techniques introduced by TD3 to learn value functions. Meanwhile, the variance involved in policy training can be noticeably reduced with the help of some variance control techniques \cite{shi2019}. For example, the \emph{stochastic variance-reduced policy gradient} (SVRPG) algorithm reuses past gradient information to reduce variance of newly estimated PGs \cite{papini2018}. All PGs are estimated through G(PO)MDP \cite{baxter2001} based on on-policy samples\footnote{In our experiments, SVRPG did not show superior performance in comparison to cutting-edge algorithms such as TD3 and SAC. Hence experiment results in relation to SVRPG will not be reported in this paper.}.

\emph{KL-divergence} is commonly used to regularize behavioral deviation of stochastic policies \cite{kaplanis2019}. Since KL-divergence does not apply to deterministic policies, we propose to measure the deviation in action spaces as a new regularizer. We also build a VAE-based generative model to extract essential policy behavioral knowledge from the elite ERB and use the model to regularize policy training. VAE has been frequently explored to support meta-RL, transfer learning and multi-task learning \cite{rakelly2019,kaplanis2018,teh2017}. We propose a new use of VAE for estimating elite DPGs in this paper.

\section{Preliminaries}
\label{sec-bg}

\subsection{Markov Decision Process and Actor-Critic Reinforcement Learning}

We are interested in MDPs with continuous action spaces. At any time $t$, an RL agent has access to the current state of its learning environment, denoted as $s_t\in\mathbb{S}$. Based on this, the agent performs an action $a_t$ selected from an $m$-dimensional continuous action space $\mathbb{A}\subset\mathbb{R}^m$. This causes an environment transition to a new state $s_{t+1}$ at time $t+1$, governed by unknown state-transition probability $\Pr(s_t,s_{t+1},a_t)$. The environment also produces a scalar and bounded reward $r(s_t,a_t)$ as its immediate feedback to the agent. Guided by a parametric deterministic policy $\pi_{\theta}:\mathbb{S}\rightarrow\mathbb{A}$ that specifies the action $a$ to be performed in any state $s$, the agent can generate a trajectory $\tau=\{(s_t^{\tau},s_{t+1}^{\tau},a_t^{\tau},r_t^{\tau})\}_{t=0}^{\infty}$ involving a sequence of consecutive state transitions over time, starting from an initial state $s_0$. The goal for RL is to identify the \emph{optimal policy parameters} $\theta^*$ that maximizes the expected long-term cumulative reward over all trajectories, as defined below:
\begin{equation}
\theta^* = \argmax_{\theta} J(\pi_{\theta})=\argmax_{\theta} \E_{\tau \sim \pi} \left[ \sum_{t=0}^{\infty} \gamma^t r(s_t^{\tau},a_t^{\tau}) \right],
\label{eq-lt-cum-rew}
\end{equation}
\noindent
with $\gamma\in(0,1)$ being a discount factor. According to the \emph{actor-critic} (AC) framework, the \emph{critic} in an AC algorithm for DRL is responsible for learning the value functions of $\pi_{\theta}$. Specifically, off-policy training of the Q-function can be performed by the critic to minimize the Bellman loss below over a batch of environment samples $\mathcal{B}^f$ collected from the full ERB:
\begin{equation}
J_{B^f}=\frac{1}{\|\mathcal{B}^f\|}\sum_{(s,s',a,r)\in\mathcal{B}^f} \left(
Q_{\omega}^{\pi_{\theta}}(s,a) - r(s,a) - \gamma \hat{Q}_{\omega}^{\pi_{\theta}}(s',\pi_{\theta}(s'))
\right)^2.
\label{equ-bell-loss}
\end{equation}
\noindent
$Q_{\omega}^{\pi_{\theta}}$ is a parametric approximation of the Q-function with trainable parameters $\omega$. $\hat{Q}_{\omega}^{\pi_{\theta}}$ in (\ref{equ-bell-loss}) represents the target Q-network \cite{fujimoto2018}. Guided by the value functions learned by the critic, the \emph{actor} in the AC algorithm proceeds to train policy $\pi_{\theta}$ by using DPGs estimated according to:
\begin{equation}
\nabla_{\theta}J(\pi_{\theta})\approx\E_{s_t\sim \rho^{\beta}} \left[ \nabla_{\theta} \pi_{\theta}(s_t) \nabla_a Q^{\pi_{\theta}}_{\omega}(s_t,a)|_{a=\pi_{\theta}(s_t)} \right],
\label{eq-dpg}
\end{equation}
\noindent
where $\beta$ refers to the \emph{stochastic behavioral policy} for environment exploration, as defined below:
\begin{equation}
\forall s\in\mathbb{S}, \beta(s)=\pi_{\theta}(s)+\mu_a,
\label{eq-sp}
\end{equation}
\noindent
with $\mu_a\sim \mathcal{N}(0,\Sigma_a)$ and $\Sigma_a$ is an $m\times m$ diagonal covariance matrix that controls the scale of environment exploration. $\rho^{\beta}$ in (\ref{eq-dpg}) stands for the \emph{discounted state visitation distribution} induced by following policy $\beta$ \cite{sutton2000}. In practice, DDPG and TD3 use environment samples stored in the full ERB to estimate $\nabla_{\theta}J(\pi_{\theta})$. Policy $\pi_{\theta}$ is hence trained with conventional DPGs alone in the two algorithms.

\subsection{Twin-Delayed Deep Deterministic Policy Gradient}

The failure of DDPG is commonly associated with overestimation of the Q-function, which may cause divergence of the actor \cite{henderson2017arxiv,fujimoto2018}. To improve learning stability of DDPG, TD3 introduces three key innovations: (1) {\bf Clipped double Q-learning}: TD3 learns two Q-functions in parallel and uses the smaller of the two to determine the target Q-value in the Bellman loss defined in (\ref{equ-bell-loss}). (2) {\bf Delayed policy update}: TD3 trains policy less frequently than Q-function. As a rule of thumb, every two updates of Q-function will be followed by one update of policy. According to \cite{fujimoto2018}, the less frequent policy updates can benefit from low-variance estimation of Q-function, thereby improving the quality of policy training. (3) {\bf Target policy smoothing}: TD3 adds noise to the target action $\pi_{\theta}(s')$ in (\ref{equ-bell-loss}) to enforce the notion that any two similar actions $a$ and $a'$ should have similar Q-values in any state $s$, thereby mitigating the impact of critic error on policy training. The added noise is also clipped to keep the target action close to the original action.

\section{Merge Deterministic Policy Gradient with Varied Bias-Variance Tradeoff}
\label{sec-meth}

In this section, we first study the techniques to estimate conventional DPG and elite DPG. Eligible methods for merging conventional DPG and elite DPG will be proposed subsequently. Finally theoretical analysis will be performed to bound the performance impact of using elite DPGs under certain conditions and assumptions. We will also show that both interpolation merging and two-step merging methods can reduce variance of trained policies with controllable bias.

\subsection{Estimate Deterministic Policy Gradient}
\label{sub-sec-est}

We study two different approaches to estimate DPG. Similar to DDPG and TD3, the first approach aims at estimating DPG with low bias but potentially high variance. For this purpose, we randomly sample a batch of records $\mathcal{B}^f$ from the full ERB and use the batch to estimate conventional DPG, denoted as $\nabla^c_{\theta}J(\pi_{\theta})$. Because these samples were obtained by using many different policies, we can effectively mitigate the possible bias induced by any specific policies. Hence, DDPG and TD3 can effectively train a policy network by using $\nabla^c_{\theta}J(\pi_{\theta})$ alone \cite{lillicrap2015,fujimoto2018}. On the other hand, since only a small collection $\mathcal{B}^f$ is sampled from thousands or millions of records stored in the full ERB\footnote{In our experiments, full ERB contains up to 1M environment samples. On the other hand, the size of $\mathcal{B}^f$ is at most 256 as recommended in some related works \cite{haarnoja2018icml}.}, the estimation of $\nabla^c_{\theta}J(\pi_{\theta})$ can vary substantially, depending on the actual samples contained in $\mathcal{B}^f$. Such high variance may noticeably slow down the learning progress.

To address this issue, we propose a new way of estimating DPG with low variance at the cost of increased bias. Guided by the aim for the newly estimated DPG to increase the chance for an RL agent to generate trajectories with high cumulative rewards, we decide to maintain a separate elite ERB and sample the \emph{elite batch}, denoted as $\mathcal{B}^e$, from the elite ERB to estimate DPG. The elite ERB always contain the top $\kappa$\footnote{We set $\kappa$ to 30 for all experiments in Section \ref{sec-ex}. Changing $\kappa$ does not seem to have major impact on algorithm performance. Given that every trajectory can contain a maximum of 1000 state transitions and the maximum size of the full ERB is 1M samples, only a very small portion of samples in the full ERB will also appear in the elite ERB.} trajectories with the highest cumulative rewards ever observed.

Using (\ref{eq-dpg}) to compute the elite DPG $\nabla^e_{\theta}J(\pi_{\theta})$ does not necessarily increase the chance for an RL agent to produce high-reward trajectories. Note that in (\ref{eq-dpg}) the action $a$ in any sampled state $s_t$ is determined completely by policy $\pi_{\theta}$ and $\pi_{\theta}(s_t)$ may deviate substantially from the sampled action $a_t$ recorded as part of the elite trajectories. Such behavioral deviation poses difficulties for the RL agent to progressively improve on its past success (i.e., elite trajectories). To tackle this problem and to further reduce the variance involved in estimating the elite DPG, we design a new regularizer based on the policy consolidation mechanism to control behavioral deviation of trained policies. Specifically, the regularized $\nabla^e_{\theta}J(\pi_{\theta})$ can be estimated as below:
\begin{equation}
\nabla^e_{\theta}J(\pi_{\theta})\approx \frac{1}{\|\mathcal{B}^e\|} \sum_{(s,s',a)\in\mathcal{B}^e}  \nabla_{\theta} \pi_{\theta}(s) \nabla_b Q_{\omega}^{\pi_{\theta}}(s,b)|_{b=\pi_{\theta}(s)} - \lambda \nabla_{\theta} \|\pi_{\theta}(s)-a\|_2^2,
\label{eq-e-dgp}
\end{equation}
\noindent
with $\|a\|_2$ representing the \emph{$l^2$-norm} of $a$ and $\lambda \geq 0$ serving as the scalar \emph{regularization factor}. There is a major issue associated with using (\ref{eq-e-dgp}). Note that any sampled action recorded in the elite ERB was generated by a stochastic policy $\beta$ defined in (\ref{eq-sp}) and is subject to a certain level of noise. Upon forcing the policy $\pi_{\theta}$ under training to follow such noisy actions, action-selection error made in any sampled state may be preserved by (\ref{eq-e-dgp}) over many learning iterations, resulting in slow performance improvement. To cope with this problem, we propose to extract policy behavioral knowledge from the elite ERB in the form of a VAE model, which can be further exploited to generate reference actions required by the regularizer. Accordingly, (\ref{eq-e-dgp}) can be re-formulated as below:
\begin{equation}
\nabla^e_{\theta}J(\pi_{\theta})\approx \frac{1}{\|\mathcal{B}^e\|} \sum_{(s,s',a)\in\mathcal{B}^e}  \nabla_{\theta} \pi_{\theta}(s) \nabla_b Q_{\omega}^{\pi_{\theta}}(s,b)|_{b=\pi_{\theta}(s)} - \lambda \nabla_{\theta} \|\pi_{\theta}(s)-\mathcal{V}(s)\|_2^2,
\label{eq-ev-dgp}
\end{equation}
\noindent
with $\mathcal{V}$ referring to the VAE model, which will be trained regularly by using random batches of samples retrieved from the elite ERB. It is possible to use any generative models other than VAE to extract policy behavioral knowledge. We choose VAE in this study because of our familiarity and its successful use in several existing DRL algorithms \cite{rakelly2019,kaplanis2018,teh2017}. By replacing sampled action $a$ in (\ref{eq-e-dgp}) with generative action $\mathcal{V}(s)$ in (\ref{eq-ev-dgp}), we can effectively reduce the long-term impact of any error embedded in sampled actions.

\subsection{Merge Deterministic Policy Gradient}
\label{sub-sec-mer}

Using the techniques developed in Subsection \ref{sub-sec-est}, we can estimate conventional DPG $\nabla^c_{\theta}J(\pi_{\theta})$ and elite DPG $\nabla^e_{\theta}J(\pi_{\theta})$ with varied bias-variance tradeoffs during each learning iteration of an AC algorithm such as TD3. Due to biased estimation of $\nabla^e_{\theta}J(\pi_{\theta})$, it is not desirable to use $\nabla^e_{\theta}J(\pi_{\theta})$ alone to train $\pi_{\theta}$. Considering the possible joint use of both $\nabla^c_{\theta}J(\pi_{\theta})$ and $\nabla^e_{\theta}J(\pi_{\theta})$, we can develop two distinct methods to merge them. We call these methods respectively the \emph{interpolation merging} and the \emph{two-step merging} methods. Interpolation merging linearly combines $\nabla^c_{\theta}J(\pi_{\theta})$ and $\nabla^e_{\theta}J(\pi_{\theta})$ with the help of a scalar \emph{weight factor} $0\leq\upsilon< 1$, as described below:
\begin{equation}
\nabla^{IM}_{\theta}J(\pi_{\theta})=(1-\upsilon) \nabla^c_{\theta}J(\pi_{\theta}) + \upsilon \nabla^e_{\theta}J(\pi_{\theta}).
\label{eq-dpg-im}
\end{equation}
\noindent
In the sequel, we will use $\nabla^{IM}_{\theta}J(\pi_{\theta})$ to denote DPG obtained through interpolation merging. Different from (\ref{eq-dpg-im}), it is possible for us to construct a two-step iterative process to merge conventional DPG and elite DPG. This is described in (\ref{eq-dpg-2s}) as follows:
\begin{equation}
\nabla^{2M}_{\theta}J(\pi_{\theta}) = (1-\upsilon) \nabla^c_{\theta}J(\pi_{\theta}) + \upsilon \nabla^e_{\theta'}J(\pi_{\theta'}),
\label{eq-dpg-2s}
\end{equation}
$$
\theta'= \theta + \alpha (1-\upsilon) \nabla^c_{\theta}J(\pi_{\theta}).
$$
\noindent
Here $0<\alpha<1$ stands for the \emph{learning rate} for policy training. Note that $\nabla^e_{\theta'}J(\pi_{\theta'})$ in (\ref{eq-dpg-2s}) merges conventional and elite DPGs in an iterative manner via $\theta'$. It is subsequently linearly merged with $\nabla^c_{\theta}J(\pi_{\theta})$ to obtain $\nabla^{2M}_{\theta}J(\pi_{\theta})$ as the outcome of two-step merging. In addition to $\nabla^{IM}_{\theta}J(\pi_{\theta})$ and $\nabla^{2M}_{\theta}J(\pi_{\theta})$, we will also consider the baseline algorithm of using $\nabla^c_{\theta}J(\pi_{\theta})$ alone to train policy $\pi_{\theta}$ in order to examine the theoretical and practical benefits of merging. Consider specifically three separate learning rules:
\begin{equation}
\begin{split}
\theta_{i+1}^c & \leftarrow \theta_i^c + \alpha \nabla^c_{\theta_i}J(\pi_{\theta_i^c}), \\
\theta_{i+1}^{IM} & \leftarrow \theta_i^{IM} + \alpha \nabla^{IM}_{\theta_i}J(\pi_{\theta_i^{IM}}), \\
\theta_{i+1}^{2M} & \leftarrow \theta_i^{2M} + \alpha \nabla^{2M}_{\theta_i}J(\pi_{\theta_i^{2M}}),
\end{split}
\label{eq-lr}
\end{equation}
\noindent
which will be used in the $i$-th learning iteration to train policy $\pi_{\theta_i}$. By studying the asymptotic trend of $\theta_{i}$ across a long sequence of learning iterations in Subsection \ref{sub-sec-ta}, we can deepen our understanding of the interpolation merging and two-step merging methods.

Driven by the learning rules in (\ref{eq-lr}) for updating $\theta^{IM}$ and $\theta^{2M}$, we develop two new DRL algorithms based on TD3, i.e., TD3-IM and TD3-2M. Algorithm \ref{alg-td} in Appendix A describes the policy training procedure of the two algorithms as well as TD3. TD3, TD3-IM and TD3-2M adopt the same method to train Q-networks. Meanwhile, only TD3-IM and TD3-2M require to maintain the elite ERB for the purpose of estimating the elite DPG. The two algorithms differ by the DPG merging methods adopted. Other parts of the three algorithms remain identical.

\subsection{Theoretical Analysis}
\label{sub-sec-ta}

In view of the biased estimation of elite DPGs, the performance impact of using elite DPGs to train policy $\pi_{\theta}$ will be studied in this subsection. Suppose that the state space $\mathbb{S}$ is discrete and finite\footnote{Any state representation in a computer system by using a fixed number of bits satisfies this requirement.}. Consider four policies $\pi_{\theta}$, $\pi_{\theta'}$, $\beta_1$ and $\beta_2$. Among the four, policy $\beta_1$ is embodied by the state-transition samples stored in the full ERB.  Policy $\beta_2$ is embodied by the state-transition samples stored in the elite ERB. Policy $\pi_{\theta'}$ is obtained by performing one-step training of policy $\pi_{\theta}$ with sufficiently small learning rate $\alpha$. Define
\begin{equation}
\begin{split}
\tilde{J}(\pi_{\theta},\beta_2) & =J(\beta_2)+\E_{\tau\sim\beta_2}\left[ A_{\omega}^{\beta_2}(s_t,\pi_{\theta}(s_t)) \right] \\
\tilde{J}(\pi_{\theta},\beta_1,\beta_2,\upsilon) & = (1-\upsilon) \tilde{J}(\pi_{\theta},\beta_1) + \upsilon \tilde{J}(\pi_{\theta},\beta_2) \\
\tilde{J}(\pi_{\theta},\pi_{\theta'},\beta_1,\beta_2,\upsilon) &= (1-\upsilon) \tilde{J}(\pi_{\theta},\beta_1) + \upsilon \tilde{J}(\pi_{\theta'},\beta_2)
\end{split}
\label{equ-j-tilde}
\end{equation}
\noindent
where $A^{\beta}_{\omega}(s_t,\pi_{\theta}(s_t))=Q_{\omega}^{\pi_{\theta}}(s_t,\pi_{\theta}(s_t))-Q_{\omega}^{\pi_{\theta}}(s_t,\beta(s_t))$. It is straightforward to see that $\nabla^e_{\theta}J(\pi_{\theta})\approx \nabla_{\theta}\tilde{J}(\pi_{\theta},\beta_2)$, $\nabla^{IM}_{\theta}J(\pi_{\theta})\approx \nabla_{\theta} \tilde{J}(\pi_{\theta},\beta_1,\beta_2,\upsilon)$ and $\nabla^{2M}_{\theta}J(\pi_{\theta})\approx \nabla_{\theta} \tilde{J}(\pi_{\theta},\pi_{\theta'},\beta_1,\beta_2,\upsilon)$. Hence, $\tilde{J}(\pi_{\theta},\beta_2)$, $\tilde{J}(\pi_{\theta},\beta_1,\beta_2,\upsilon)$ and $\tilde{J}(\pi_{\theta},\pi_{\theta'},\beta_1,\beta_2,\upsilon)$ enable us to bound the bias involved in training policy $\pi_{\theta}$ by using $\nabla^e_{\theta}J(\pi_{\theta})$, $\nabla^{IM}_{\theta}J(\pi_{\theta})$ and $\nabla^{2M}_{\theta}J(\pi_{\theta})$ respectively. Assume that, for any two actions $a,a'\in\mathbb{A}$, whenever their difference $\|a-a'\|_1$ is sufficiently small, there exists a positive constant $\chi_1$ such that $\max_{s\in\mathbb{S}} |Q_{\omega}^{\pi_{\theta}}(s,a)-Q_{\omega}^{\pi_{\theta}}(s,a')|\leq\chi_1 \|a-a_1\|_1$. Further assume that, for sufficiently small $\alpha$, there exists a positive constant $\chi_2$ such that $\max_{s\in\mathbb{S}} \|\pi_{\theta}(s)-\pi_{\theta'}(s) \|_1\leq \chi_2\alpha$. Define $\zeta_{\pi_{\theta}}=\max_{s\in\mathbb{S}} |Q_{\omega}^{\pi_{\theta}}(s,\pi_{\theta}(s))|<\infty$. Proposition \ref{po-perf-bound} below gives the bound on the learning bias associated with $\tilde{J}(\pi_{\theta},\beta_2)$, $\tilde{J}(\pi_{\theta},\beta_1,\beta_2,\upsilon)$ and $\tilde{J}(\pi_{\theta},\pi_{\theta'},\beta_1,\beta_2,\upsilon)$. See Appendix B for proof.
\begin{proposition}
Regarding $\tilde{J}(\pi_{\theta},\beta_2)$, $\tilde{J}(\pi_{\theta},\beta_1,\beta_2,\upsilon)$ and $\tilde{J}(\pi_{\theta},\pi_{\theta'},\beta_1,\beta_2,\upsilon)$ defined in (\ref{equ-j-tilde}), their differences from $J(\pi_{\theta})$ are bounded as below:
\begin{equation}
\begin{split}
\left\| J(\pi_{\theta}) - \tilde{J}(\pi_{\theta},\beta_2) \right\|_1 & \leq 2 \zeta_{\pi_{\theta}} D_{TV} (\rho_0^{\pi_{\theta}}, \rho_0^{\beta_2}), \\
\left\| J(\pi_{\theta}) - \tilde{J}(\pi_{\theta},\beta_1,\beta_2,\upsilon) \right\|_1 & \leq 2 \zeta_{\pi_{\theta}} \upsilon D_{TV}(\rho_0^{\pi_{\theta}},\rho_0^{\beta_2}), \\
\left\| J(\pi_{\theta}) - \tilde{J}(\pi_{\theta},\pi_{\theta'},\beta_1,\beta_2,\upsilon) \right\|_1 & \leq 2 \zeta_{\pi_{\theta}} \upsilon D_{TV}(\rho_0^{\pi_{\theta}},\rho_0^{\beta_2}) + \frac{\upsilon\chi_1\chi_2\alpha}{1-\gamma}
\end{split}
\label{equ-j-bound}
\end{equation}
\noindent
where $D_{TV}$ refers to the total variation distance between $\rho_0^{\pi_{\theta}}$ and $\rho_0^{\beta_2}$ \cite{cha2007}.
\label{po-perf-bound}
\end{proposition}
$D_{TV}(\rho^{\pi_{\theta}}_0, \rho^{\beta}_0)$ measures the probability differences between $\pi_{\theta}$ and $\beta_2$ in selecting the initial state $s_0$. It can be reduced by increasing the size of the elite ERB. Upon using the full ERB to estimate conventional DPG $\nabla^c_{\theta}J(\pi_{\theta})$, $D_{TV}(\rho^{\pi_{\theta}}_0, \rho^{\beta_1}_0)\approx 0$. This is because the distribution of initial states recorded in the full ERB is very close to $\rho_0^{\pi_{\theta}}$, which is determined completely by the learning environment. Moreover, (\ref{equ-j-bound}) indicates that the bias of $\tilde{J}(\pi_{\theta},\beta_1,\beta_2,\upsilon)$ is less than that of $\tilde{J}(\pi_{\theta},\beta_2)$ since $\upsilon<1$. Upon using small $\alpha$, $\tilde{J}(\pi_{\theta},\pi_{\theta'},\beta_1,\beta_2,\upsilon)$ can also keep its bias at a low level. As explained in Appendix B, theoretically Proposition \ref{po-perf-bound} produces three new learning rules with guaranteed monotonic policy improvements, which can be realized approximately and efficiently by training policy $\pi_{\theta}$ via $\nabla_{\theta}^e J(\pi_{\theta})$, $\nabla_{\theta}^{IM}J(\pi_{\theta})$ and $\nabla_{\theta}^{2M}J(\pi_{\theta})$. Besides Proposition \ref{po-perf-bound}, additional analysis has also be performed in Appendix C to study the bias introduced by imprecise approximations of the value function and to bound directly the bias involved in estimating and merging DPGs. The analysis provides theoretical clues regarding the importance of adopting the regularizer in (\ref{eq-e-dgp}) and (\ref{eq-ev-dgp}). To further analyze the bias-variance tradeoffs achievable by using $\nabla_{\theta}^{IM}J(\pi_{\theta})$ and $\nabla_{\theta}^{2M}J(\pi_{\theta})$, we perform a \emph{noisy quadratic analysis} below \cite{zhang2019}. Assuming that the policy parameters $\theta$ under training are close to the optimal policy parameters $\theta^*$, $J(\pi_{\theta})$ and $\tilde{J}(\pi_{\theta},\beta_2)$ can be approximated respectively by the following \emph{noisy quadratic models}:
\begin{equation}
J(\pi_{\theta})\approx J^* - \frac{1}{2} (\theta-c_1)^T A (\theta-c_1), \tilde{J}(\pi_{\theta},\beta_2)\approx J^* - \frac{1}{2} (\theta-c_2)^T A (\theta-c_2).
\label{eq-nqm}
\end{equation}
\noindent
Assume without loss of generality that $\theta^*=0$, similar to \cite{wu2018,schaul2013}. In (\ref{eq-nqm}), $c_1\sim \mathcal{N}(0,\Sigma_1)$\footnote{The analysis can be easily expanded to include the bias involved in estimating $\nabla^c_{\theta}J(\pi_{\theta})$. However such bias is ignored in this paper since it is expected to be much smaller than the estimation bias of $\nabla^e_{\theta}J(\pi_{\theta})$.} and $c_2\sim \mathcal{N}(\epsilon,\Sigma_2)$ with $\epsilon\neq 0$ that reflects the estimation bias associated with $\nabla^e_{\theta}J(\pi_{\theta})$. Both $\Sigma_1$ and $\Sigma_2$ are diagonal matrices with positive diagonal elements and $Diag(\Sigma_1)\gg Diag(\Sigma_2)$. In line with (\ref{eq-nqm}), $\nabla^c_{\theta}J(\pi_{\theta})\approx -A(\theta-c_1)$ and $\nabla^e_{\theta}J(\pi_{\theta})\approx \nabla_{\theta} \tilde{J}(\pi,\beta_2)\approx -A(\theta-c_2)$, where the diagonal matrix $A$ with positive diagonal elements controls the sensitivity of $J(\pi_{\theta})$ with respect to $\theta$. Despite of being simple, (\ref{eq-nqm}) remains as challenging optimization targets in the literature \cite{zhang2019,schaul2013}. Our noisy quadratic analysis produces Proposition \ref{po-nqa} below. Its proof is presented in Appendix D.
\begin{proposition}
Based on the noisy quadratic models in (\ref{eq-nqm}) and provided that $\alpha$ is sufficiently small such that $\alpha Diag(A) < Diag(I)$, policy parameters trained by using the three learning rules in (\ref{eq-lr}) have the following convergence properties in expectation:
\begin{equation}
\begin{split}
& \lim_{i\rightarrow\infty} \E[\theta_i^c]=0, \lim_{i\rightarrow\infty} \E[\theta_i^{IM}]=\upsilon \epsilon \\
& \lim_{i\rightarrow\infty} \E[\theta_i^{2M}]=\alpha \upsilon \left[ I- (I-\alpha\upsilon A) (I-\alpha (1-\upsilon) A) \right]^{-1} A \epsilon.
\end{split}
\label{equ-conv-exp}
\end{equation}
\noindent
Moreover, the variances of trained policy parameters converge to the following fixed points:
\begin{equation*}
\begin{split}
& \lim_{i\rightarrow\infty} \V[\theta_i^c]=\alpha^2 \left[ I-(I-\alpha A)^2 \right]^{-1} A^2 \Sigma_1, \\
& \lim_{i\rightarrow\infty} \V[\theta_i^{IM}] = \alpha^2 \left[ I-(I-\alpha A)^2 \right]^{-1}  A^2 \left( (1-\upsilon)^2 \Sigma_1 + \upsilon^2 \Sigma_2 \right), \\
& \lim_{i\rightarrow\infty} \V[\theta_i^{2M}]= \alpha^2 \left[ I- (I-\alpha\upsilon A)^2 (I-\alpha(1-\upsilon)A)^2 \right]^{-1} A^2 \left( (1-\upsilon)^2 (I-\alpha\upsilon A)^2\Sigma_1 + \upsilon^2 \Sigma_2 \right).
\end{split}
\end{equation*}
\label{po-nqa}
\end{proposition}
Consistent with Proposition \ref{po-perf-bound}, Proposition \ref{po-nqa} indicates that the bias involved training $\theta^{IM}$ and $\theta^{2M}$ according to (\ref{eq-lr}) can be controlled via $\upsilon$. Meanwhile, $\theta^{IM}$ enjoys less bias than $\theta^{2M}$ at the expense of relatively higher variances. In fact, because $Diag(\Sigma_2)\ll Diag(\Sigma_1)$, it can be easily verified that $\V[\theta_i^{IM}]> \V[\theta_i^{2M}]$ when $0<\upsilon<1$. Meanwhile $\V[\theta_i^{IM}]<\V[\theta_i^c]$. Therefore the three learning rules in (\ref{eq-lr}) provide varied bias-variance tradeoffs while training policy $\pi_{\theta}$.

\section{Experiments}
\label{sec-ex}

Experiments have been performed on six benchmark control tasks, including Ant, Half Cheetah, Hopper, Lunar Lander, Walker2D and Bipedal Walker. We adopt a popular implementation of these benchmarks provided by OpenAI GYM\footnote{https://gym.openai.com} and powered by the PyBullet physics engine \cite{tan2018}. Many previous studies utilized a different version of these benchmarks that rely on the MuJoCo physics engine with varied system dynamics and reward schemes \cite{todorov2012}. We prefer to use the PyBullet version since the PyBullet software package is publicly available with increasing popularity. PyBullet benchmarks are also widely considered to be tougher to solve than their MuJoCo counterparts \cite{tan2018}.

\begin{figure*}[!ht]
\begin{minipage}[t]{0.33\textwidth}
\includegraphics[width=\textwidth]{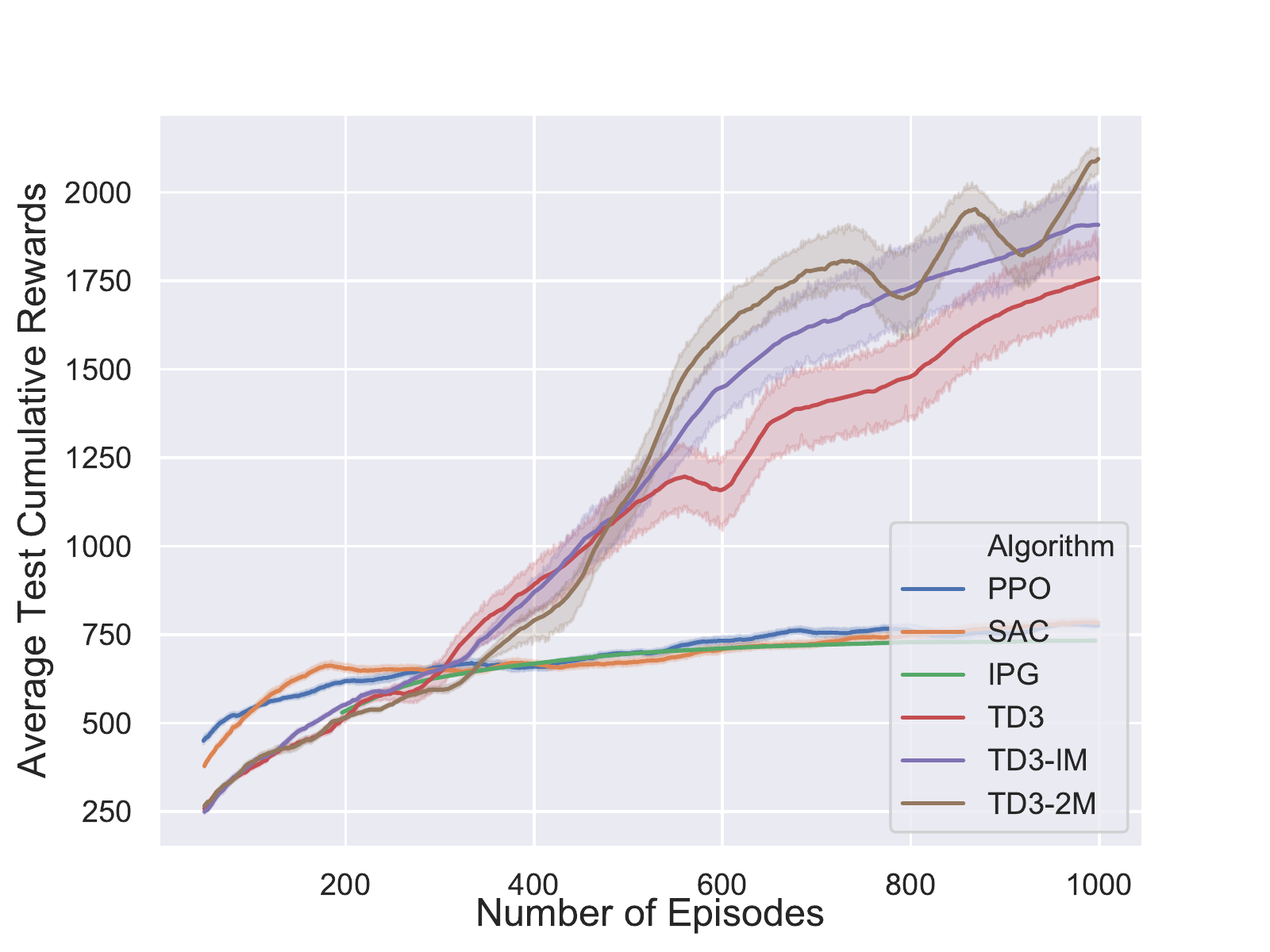}
\subcaption{Ant}
\end{minipage}
\begin{minipage}[t]{0.33\textwidth}
\includegraphics[width=\textwidth]{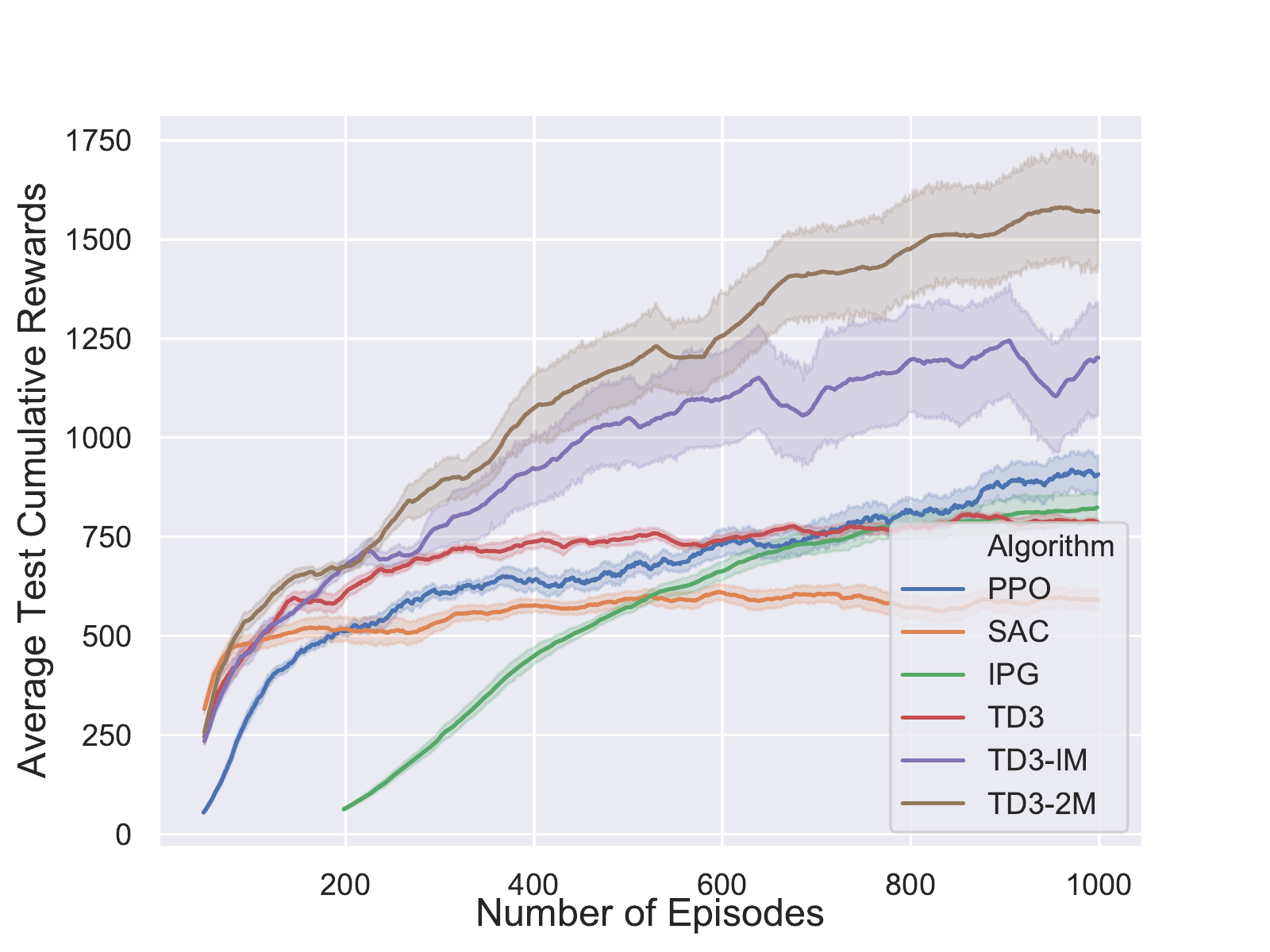}
\subcaption{Half Cheetah}
\end{minipage}
\begin{minipage}[t]{0.33\textwidth}
\includegraphics[width=\textwidth]{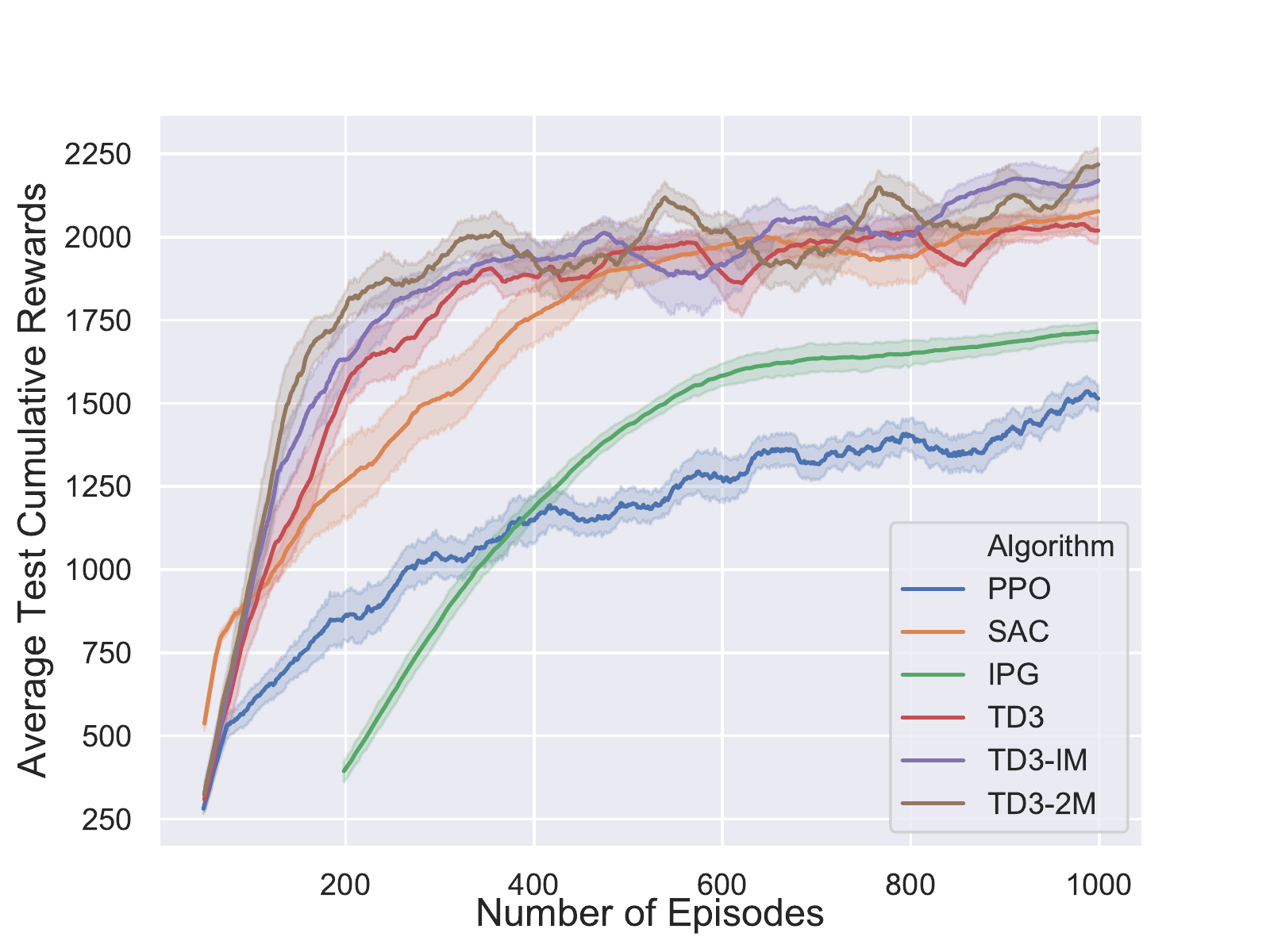}
\subcaption{Hopper}
\end{minipage}
\\
\begin{minipage}[t]{0.33\textwidth}
\includegraphics[width=\textwidth]{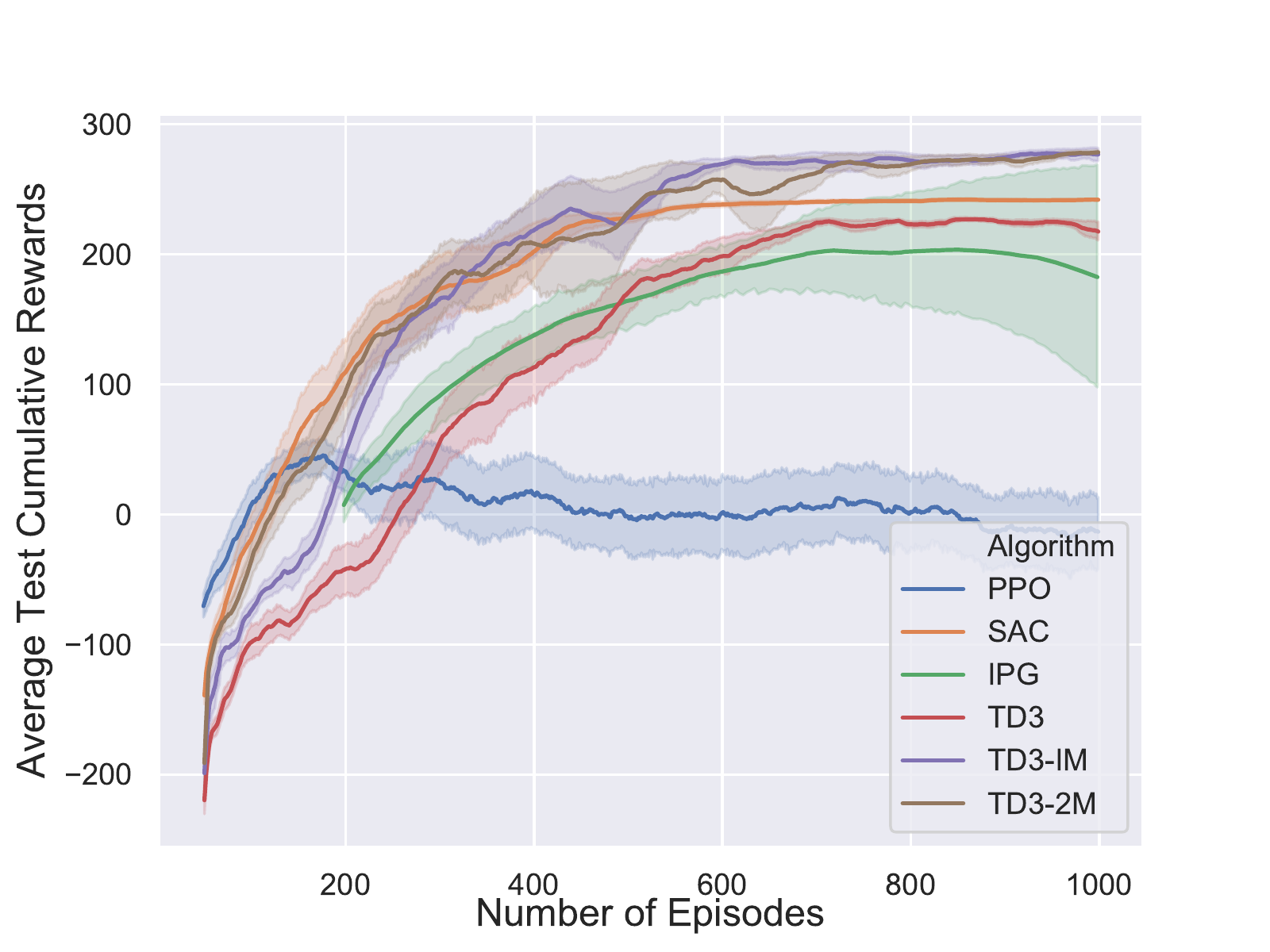}
\subcaption{Lunar Lander}
\end{minipage}
\begin{minipage}[t]{0.33\textwidth}
\includegraphics[width=\textwidth]{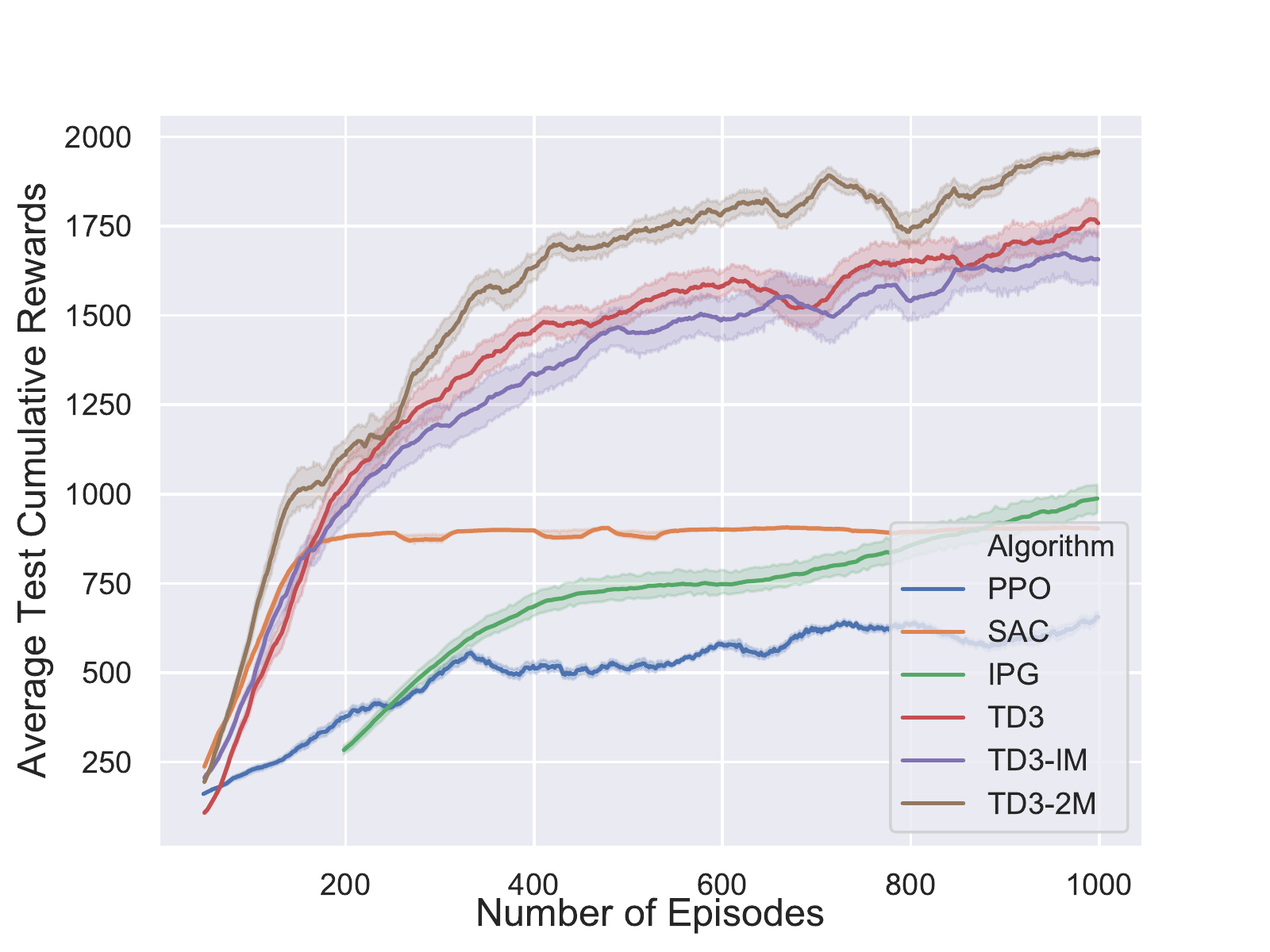}
\subcaption{Walker2D}
\end{minipage}
\begin{minipage}[t]{0.33\textwidth}
\includegraphics[width=\textwidth]{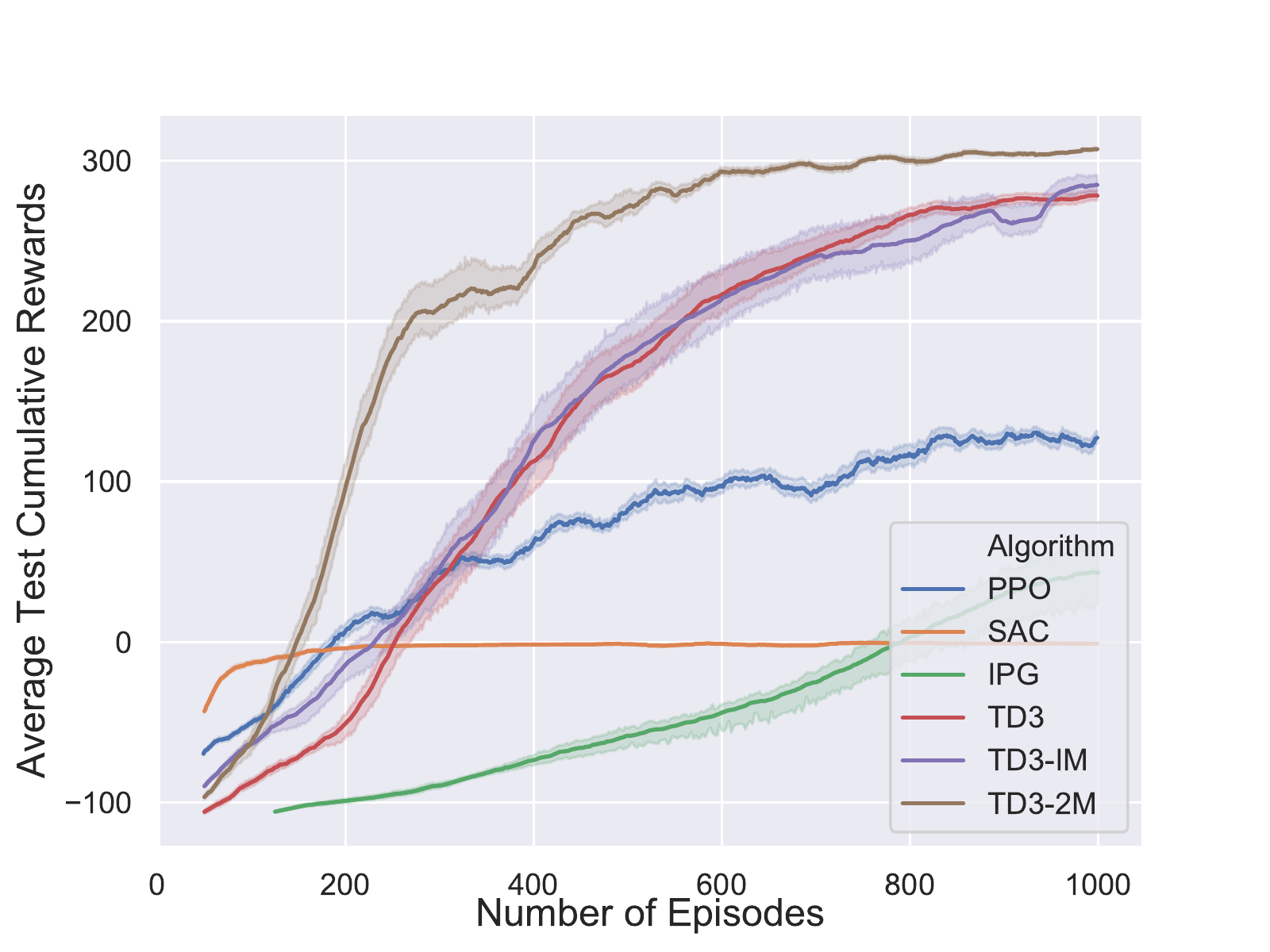}
\subcaption{Bipedal Walker}
\end{minipage}
\caption{Learning performance of PPO, SAC, IPG, TD3, TD3-IM and TD3-2M on six benchmark control problems.}
\label{fig-perf-comp}
\end{figure*}

There are six competing algorithms in our experiments. Four of them, i.e., IPG, TD3, SAC and PPO, are cutting-edge DRL algorithms with leading performance on many continuous action benchmarks. Our experiments are performed on the high-quality open source implementations of TD3, SAC and PPO provided by OpenAI Spinning Up. The TD3 code is also expanded to implement TD3-IM and TD3-2M. Moreover, we adopt the reference implementation of IPG published by its inventors. In our experiments, each learning episode contains up to 1000 sampled state transitions. A learning algorithm can learn through 1000 episodes to find the best possible policy. To evaluate the learning performance reliably, 10 independent tests of each learning algorithm have been performed on every benchmark with 10 random seeds. Detailed hyper-parameter settings of all competing algorithms can be found in Appendix E.

\begin{figure*}[!ht]
\center{
\begin{minipage}[t]{0.36\textwidth}
\includegraphics[width=\textwidth]{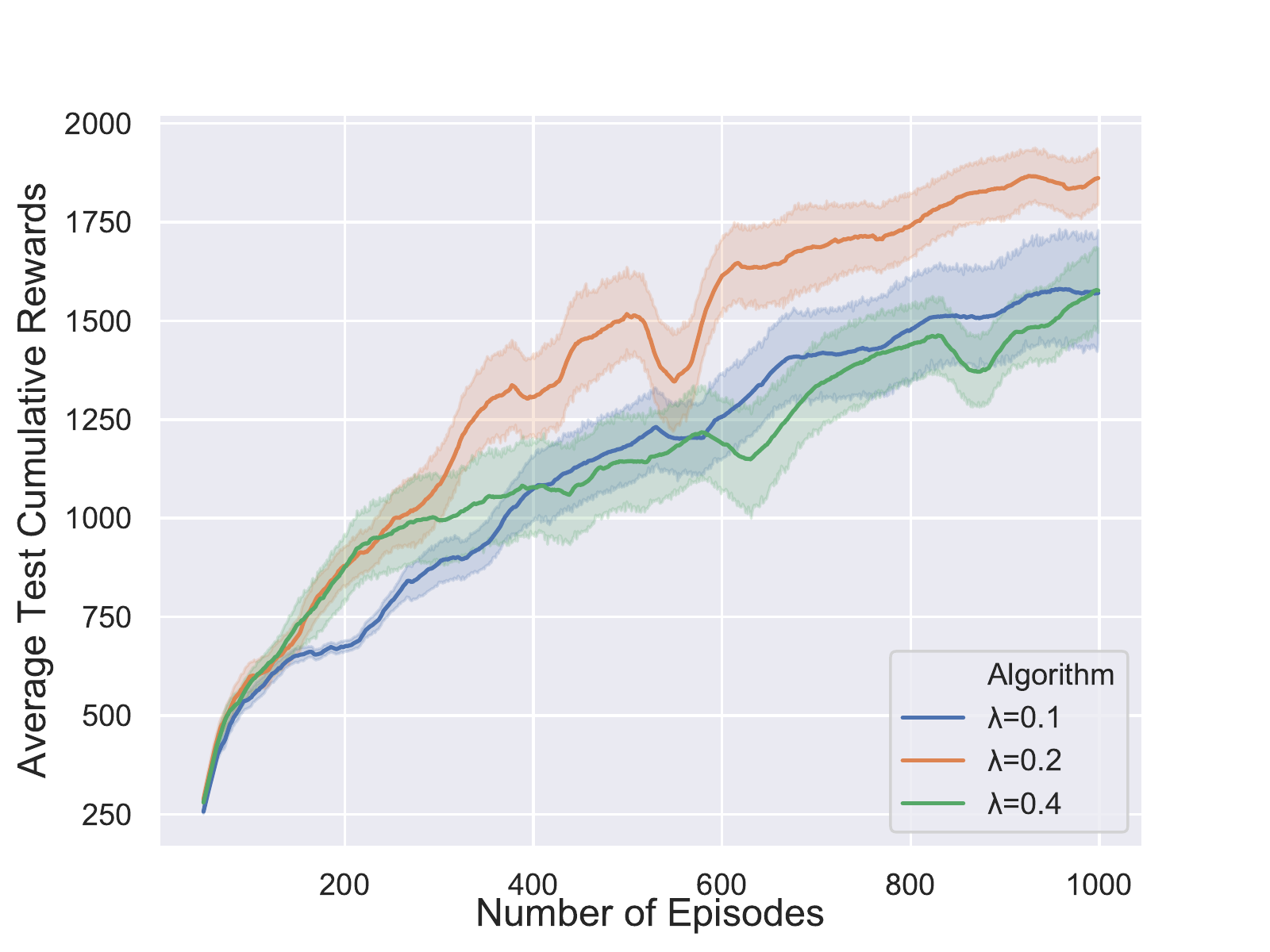}
\subcaption{Performance impact of $\lambda$.}
\end{minipage}
\hspace{0.2cm}
\begin{minipage}[t]{0.36\textwidth}
\includegraphics[width=\textwidth]{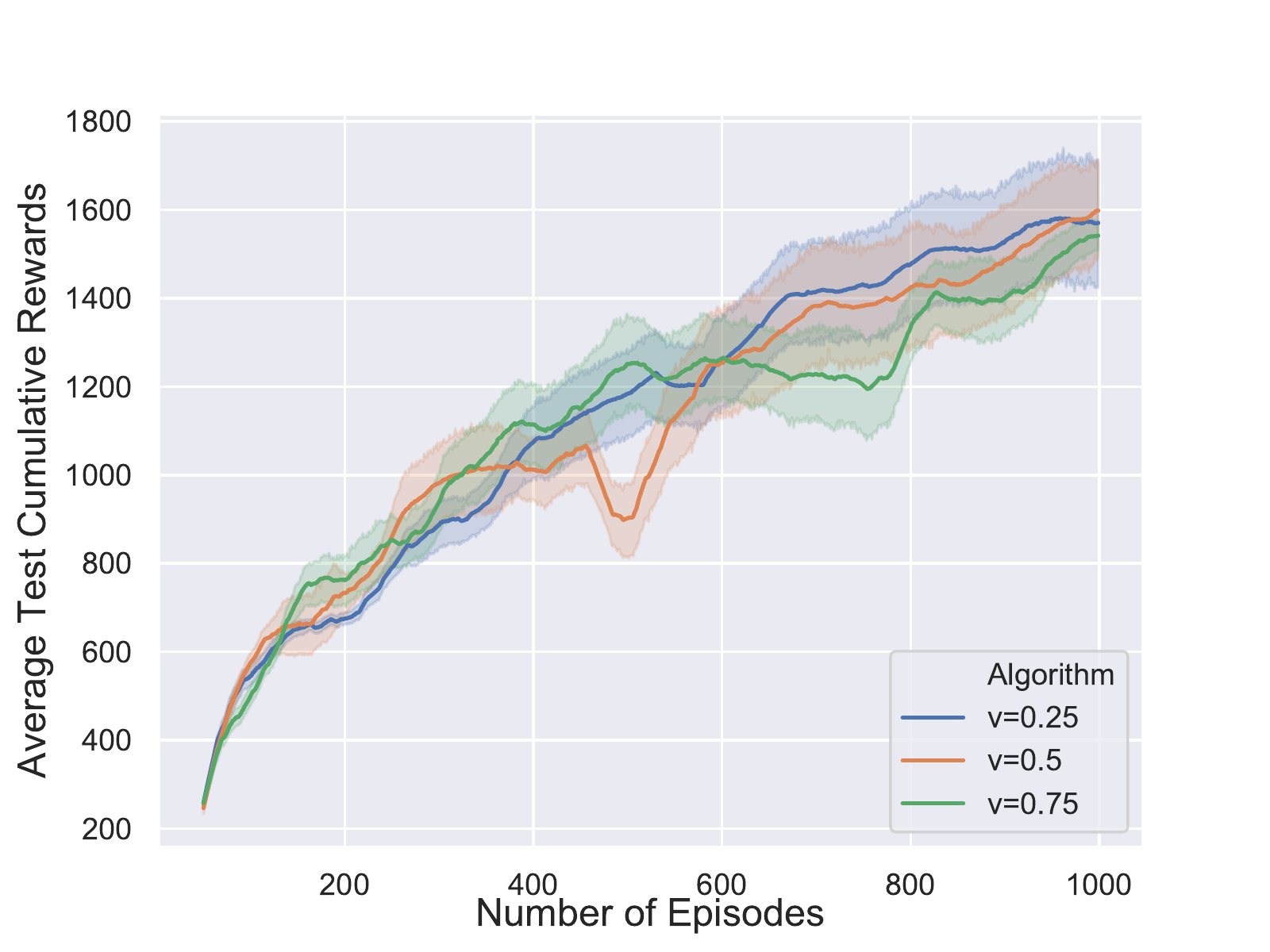}
\subcaption{Performance impact of $\upsilon$.}
\end{minipage}
}
\caption{The influence of the regularization factor $\lambda$ and the weight factor $\upsilon$ on the learning performance of TD3-2M on the Half Cheetah benchmark.}
\label{fig-para-comp}
\end{figure*}

\begin{figure*}[!ht]
\center{
\begin{minipage}[t]{0.36\textwidth}
\includegraphics[width=\textwidth]{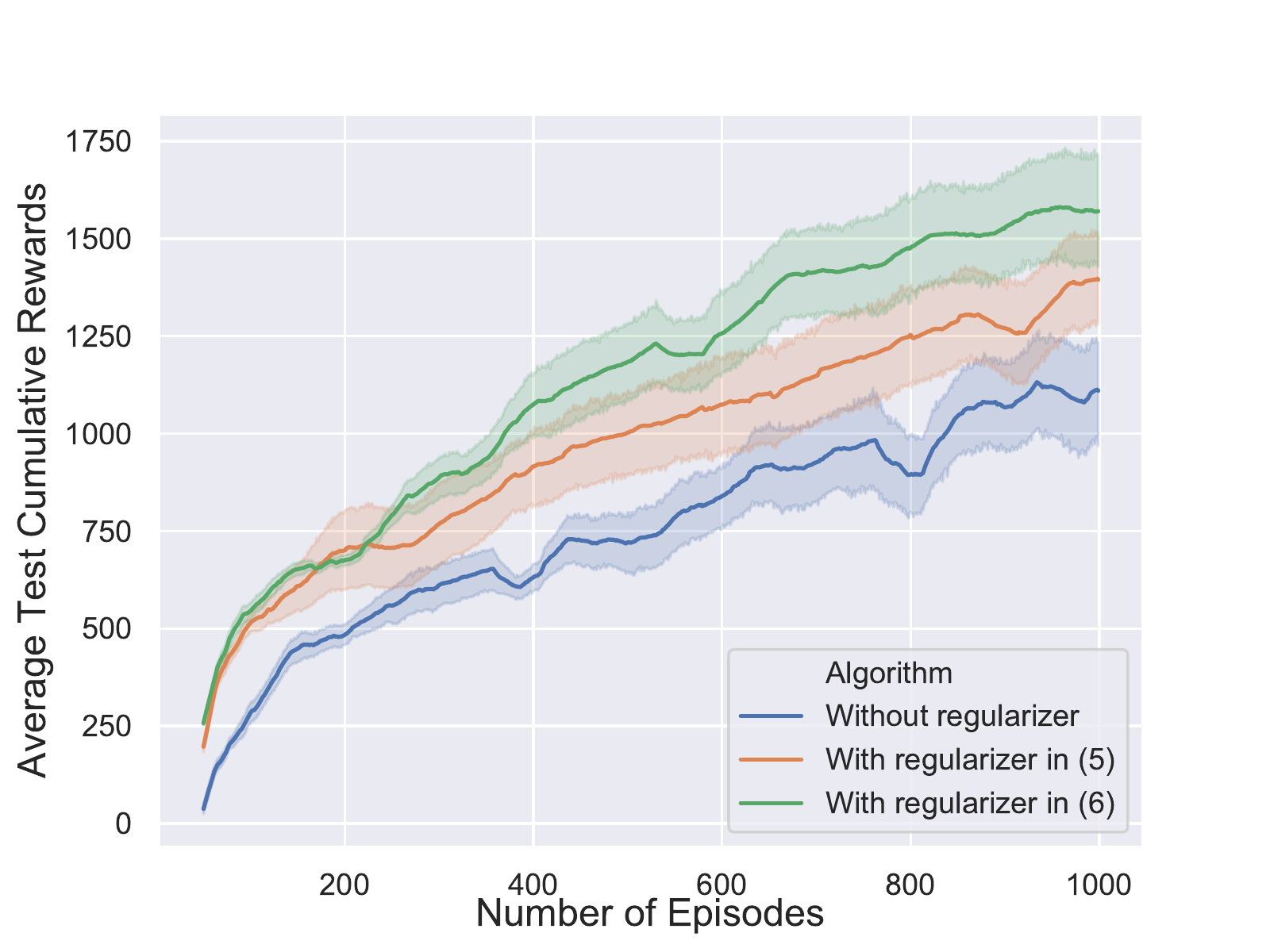}
\subcaption{Performance difference of TD3-2M upon using (\ref{eq-e-dgp}), (\ref{eq-ev-dgp}) and ``no regularizer" to estimate elite DPGs.}
\end{minipage}
\hspace{0.2cm}
\begin{minipage}[t]{0.36\textwidth}
\includegraphics[width=\textwidth]{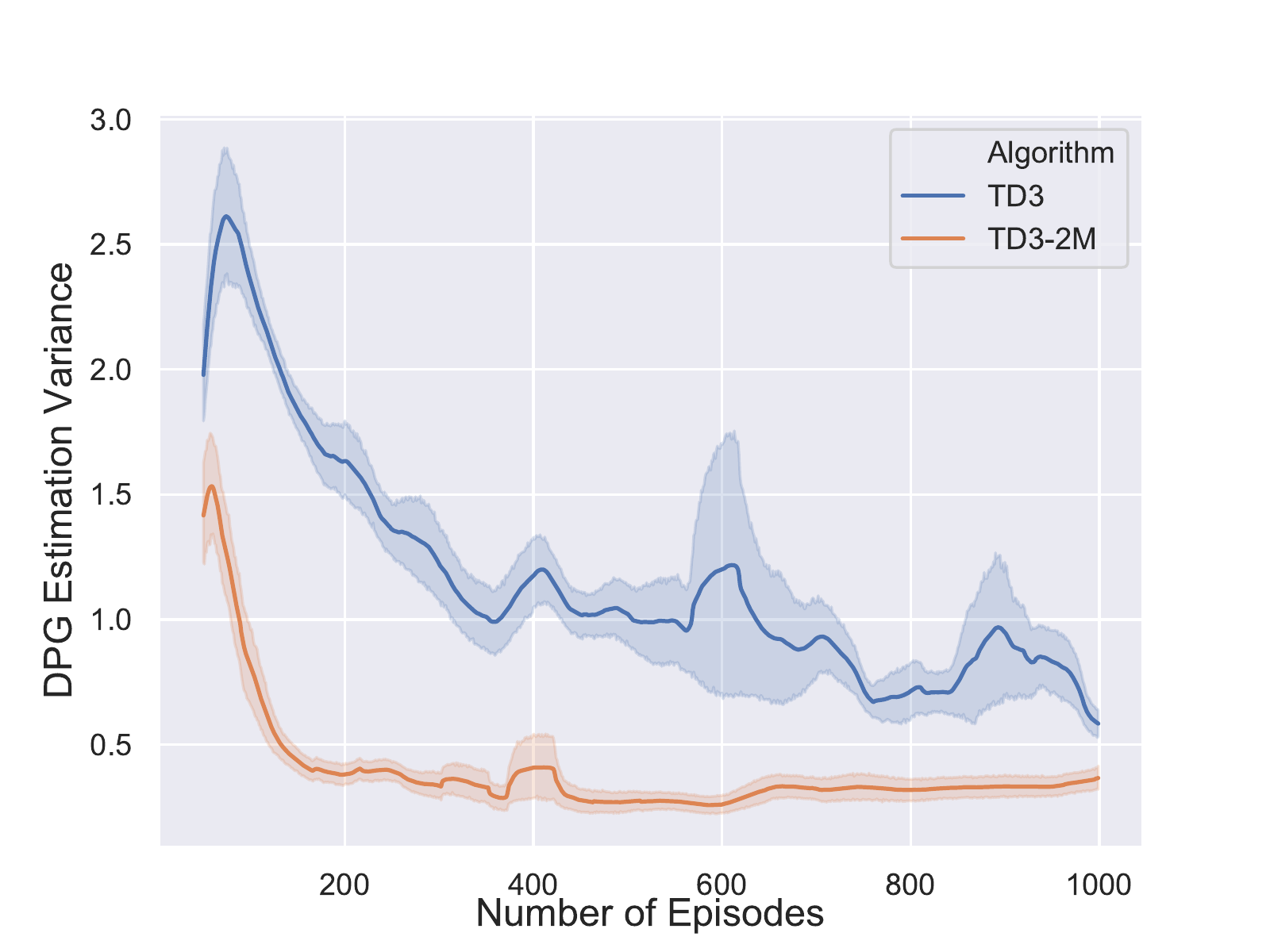}
\subcaption{The difference in DPG estimation variance between TD3 and TD3-2M.}
\end{minipage}
}
\caption{Further analysis of learning performance of TD3-2M on the Half Cheetah benchmark.}
\label{fig-reg-comp}
\end{figure*}

Figure \ref{fig-perf-comp} compares the learning performance of all algorithms on six benchmark problems. As evidenced, TD3-2M achieved consistently the best performance on all benchmarks. Specifically, on Half Cheetah, Walker2D and Bipedal Walker, the final performance of TD3-2M is much better than other algorithms. On problems such as Ant and Lunar Lander, TD3-2M achieved similar performance as TD3-IM and outperformed the rest. On Hopper, the two new algorithms also managed to achieve slightly better performance than SAC and TD3. Since TD3-IM and TD3-2M achieved overall better performance than TD3, it can be confirmed that merging conventional and elite DPGs can improve policy training effectiveness. The fact that TD3-2M performed more effectively than TD3-IM also verifies the usefulness of the newly developed two-step merging method. Besides TD3-IM, interpolation merging has also been exploited to merge on-policy PG with off-policy PG in IPG \cite{gu2017}. Despite of using the same merging method, TD3-IM can significantly outperform IPG on most of the benchmarks.

In Figure \ref{fig-perf-comp}, the performance of TD3-IM and TD3-2M was obtained with $\upsilon=0.25$ and $\lambda=0.1$. We also examined other settings of the two hyper-parameters. As demonstrated in Figure \ref{fig-para-comp}, on Half Cheetah, the performance of TD3-2M is not heavily influenced by $\upsilon$. Meanwhile, $\lambda$ seems to have stronger influence than $\upsilon$. With $\lambda=0.2$, TD3-2M achieved clearly better performance than other settings of $\lambda$. However, setting $\lambda$ to 0.1 appears to produce more reliable learning behavior. While we reported only the influence of $\lambda$ and $\upsilon$ on Half Cheetah, similar observations have been witnessed on other benchmarks. It is also suitable for TD3-IM to adopt identical settings of $\upsilon$ and $\lambda$ as TD3-2M.

Figure \ref{fig-reg-comp}(a) compares the performance of TD3-2M upon using either (\ref{eq-e-dgp}) or (\ref{eq-ev-dgp}) to estimate elite DPGs. Our experiments clearly show the effectiveness of using (\ref{eq-ev-dgp}). In particular, with (\ref{eq-ev-dgp}), steady performance improvement can be realized throughout the entire learning process. In comparison, learning slows down noticeably after the initial phase when elite DPG is estimated through (\ref{eq-e-dgp}) or without using the regularizer term in (\ref{eq-ev-dgp}). Figure \ref{fig-reg-comp}(b) further shows that TD3-2M can substantially reduce the variance of estimated DPGs for policy training than TD3, explaining empirically why it can outperform TD3 on most of the benchmark problems.

\section{Conclusion}

In this paper we studied an important research question on how to estimate and merge DPGs with varied bias-variance tradeoffs for effective DRL. Driven by this question, we introduced the elite DPG to be estimated by using both the elitism and policy consolidation techniques. Meanwhile, we have theoretically and experimentally studied two DPG merging methods, i.e., interpolation merging and two-step merging. Both merging methods can be adopted by TD3 to balance the bias-variance tradeoffs during policy training. They noticeably improved the learning performance of TD3, significantly outperforming several state-of-the-art DRL algorithms. Additionally, we showed analytically that the performance impact of using biased estimation of elite DPGs can be effectively bounded and also mitigated through DPG merging.

\small{
\bibliographystyle{plain}
\bibliography{citefile}
}

\section*{Appendix A}

This appendix presents the pseudo code of TD3, TD3-IM and TD3-2M, as shown in Algorithm \ref{alg-td}.

\begin{algorithm}[!ht]
 \begin{algorithmic}
 \STATE {\bf Input}: Initial policy network $\pi_{\theta}$, Q-network $Q_{\omega}$, and the corresponding target networks, the VAE generative model, full ERB and elite ERB
 \STATE {\bf repeat}
 \STATE \ \ \ \ Observe current state $s$ and sample a noisy action $a$ according to policy $\beta(s)$ in (\ref{eq-sp}).
 \STATE \ \ \ \ Execute action $a$ in the environment.
 \STATE \ \ \ \ Observe state transition $(s,s',a,r)$ and keep the sampled state transition in the full ERB.
 \STATE \ \ \ \ {\bf if} end of trajectory is reached:
 \STATE \ \ \ \ \ \ \ \ Update the elite ERB and reset the environment.
 \STATE \ \ \ \ {\bf if} time to update:
 \STATE \ \ \ \ \ \ \ \  {\bf for} $j$ in range(number of updates):
 \STATE \ \ \ \ \ \ \ \ \ \ \ \ Sample batch $\mathcal{B}^f$ from full ERB.
 \STATE \ \ \ \ \ \ \ \ \ \ \ \ Sample batch $\mathcal{B}^e$ from elite ERB.
 \STATE \ \ \ \ \ \ \ \ \ \ \ \ Use TD3 to train $Q_{\omega}$ based on $\mathcal{B}^f$.
 \STATE \ \ \ \ \ \ \ \ \ \ \ \ Train the VAE generative model for TD3-IM and TD3-2M based on $\mathcal{B}^e$.
 \STATE \ \ \ \ \ \ \ \ \ \ \ \ {\bf if} $j \% \text{policy\_delay} = 0$:
 \STATE \ \ \ \ \ \ \ \ \ \ \ \ \ \ \ \ Estimate $\nabla^c_{\theta}J(\pi_{\theta})$ by using $\mathcal{B}^f$.
 \STATE \ \ \ \ \ \ \ \ \ \ \ \ \ \ \ \ {\bf TD3}: Update $\theta^c$ according to (\ref{eq-lr}).
 \STATE \ \ \ \ \ \ \ \ \ \ \ \ \ \ \ \ {\bf TD3-IM}: Estimate $\nabla^e_{\theta}J(\pi_{\theta})$ by using $\mathcal{B}^e$ and merge  $\nabla^c_{\theta}J(\pi_{\theta})$ and $\nabla^e_{\theta}J(\pi_{\theta})$ via (\ref{eq-dpg-im}). \\
 \ \ \ \ \ \ \ \ \ \ \ \ \ \ \ \ \ \ \ \ \ \ \ \ Update $\theta^{IM}$ according to (\ref{eq-lr}).
 \STATE \ \ \ \ \ \ \ \ \ \ \ \ \ \ \ \ {\bf TD3-2M}: Compute $\theta'$ according to (\ref{eq-dpg-2s}) and estimate $\nabla^e_{\theta'}J(\pi_{\theta'})$ by using $\mathcal{B}^e$. \\
\ \ \ \ \ \ \ \ \ \ \ \ \ \ \ \ \ \ \ \ \ \ \ \ Merge  $\nabla^c_{\theta}J(\pi_{\theta})$ and $\nabla^e_{\theta'}J(\pi_{\theta'})$ via (\ref{eq-dpg-2s}) and update $\theta^{2M}$ according to (\ref{eq-lr}).
 \STATE \ \ \ \ \ \ \ \ \ \ \ \ \ \ \ \ Update target policy networks and target Q-networks.
 \end{algorithmic}
\caption{The twin-delayed DDPG algorithm enhanced by merging conventional DPG and elite DPG.}
\label{alg-td}
\end{algorithm}

\section*{Appendix B}

To prove Proposition \ref{po-perf-bound}, we need to utilize Lemmas \ref{lemma-1}, which will be introduced and proved first.

\begin{lemma}
Given any policy $\beta$, we have
$$
\E_{s\sim\rho_0^{\beta}} \left[ Q_{\omega}^{\pi} (s,\pi(s)) \right]=\tilde{J}(\pi,\beta)=J(\beta)+\E_{\tau\sim\beta} \left[ A_{\omega}^{\beta}(s_t,\pi(s_t)) \right]
$$
with $A_{\omega}^{\beta}(s_t,\pi(s_t))=Q_{\omega}^{\pi}(s_t,\pi(s_t))-Q_{\omega}^{\pi}(s_t,\beta(s_t))$.
\label{lemma-1}
\end{lemma}

{\bf Proof}: This lemma can be proved by expanding $\E_{\tau\sim\beta} \left[ A_{\omega}^{\beta}(s_t,\pi(s_t)) \right]$. Particularly,
\begin{equation*}
\begin{split}
& \ \ \ \E_{\tau\sim\beta} \left[ A_{\omega}^{\beta}(s_t,\pi(s_t)) \right] \\
& = -\E_{\tau\sim\beta} \left[ \sum_{t=0}^{\infty} \gamma^t \left( Q_{\omega}^{\pi}(s_t,\beta(s_t)) - Q_{\omega}^{\pi}(s_t,\pi(s_t)) \right) \right] \\
& = -\E_{\tau\sim\beta} \left[ \sum_{t=0}^{\infty} \gamma^t \left( r(s_t,\beta(s_t)) + \gamma Q_{\omega}^{\pi} (s_{t+1},\pi(s_{t+1})) - Q_{\omega}^{\pi}(s_t,\pi(s_t)) \right) \right] \\
& = -\E_{\tau\sim\beta} \left[ \sum_{t=0}^{\infty} \gamma^t r(s_t,\beta(s_t)) \right] + \E_{\tau\sim\beta} \left[ Q_{\omega}^{\pi} (s,\pi(s)) \right] \\
& = -J(\beta) + \E_{s\sim\rho_0^{\beta}} \left[ Q_{\omega}^{\pi} (s,\pi(s)) \right]
\end{split}
\end{equation*}
This proves Lemma \ref{lemma-1}. \QEDB

Based on Lemma \ref{lemma-1}, we can proceed to prove Proposition \ref{po-perf-bound}.

{\bf Proof}: Because
$$
\tilde{J}(\pi,\beta_2)=\E_{s\sim\rho_0^{\beta_2}} \left[ Q_{\omega}^{\pi} (s_0,\pi(s)) \right],
$$
\begin{equation*}
\begin{split}
& \ \ \ \left\| J(\pi) - \tilde{J}(\pi,\beta_2) \right\|_1 \\
& = \left\| \E_{s\sim\rho_0^{\pi}} \left[ Q_{\omega}^{\pi} (s,\pi(s)) \right] - \E_{s\sim\rho_0^{\beta_2}} \left[ Q_{\omega}^{\pi} (s,\pi(s)) \right] \right\|_1 \\
& = \left\| \sum_{s\in\mathbb{S}} \left(\rho_0^{\pi}(s) - \rho_0^{\beta_2}(s) \right) Q_{\omega}^{\pi} (s,\pi(s)) \right\|_1 \\
& \leq \sum_{s\in\mathbb{S}} \left| \rho_0^{\pi}(s) - \rho_0^{\beta_2}(s) \right| \cdot \left| Q_{\omega}^{\pi} (s,\pi(s)) \right| \\
& \leq 2 \zeta_{\pi} D_{TV} (\rho_0^{\pi}, \rho_0^{\beta_2} ).
\end{split}
\end{equation*}

Similarly, it is not difficult to see from Lemma \ref{lemma-1} that
$$
\tilde{J}(\pi,\beta_1,\beta_2,\upsilon) = (1-\upsilon)\E_{s\sim\rho_0^{\beta_1}} \left[ Q_{\omega}^{\pi} (s,\pi(s)) \right] + \upsilon \E_{s\sim\rho_0^{\beta_2}} \left[ Q_{\omega}^{\pi} (s,\pi(s)) \right].
$$
Hence,
\begin{equation*}
\begin{split}
& \ \ \ \left\| J(\pi) - \tilde{J}(\pi,\beta_1,\beta_2,\upsilon) \right\|_1 \\
& = \left\| \E_{s\sim\rho_0^{\pi}} \left[ Q_{\omega}^{\pi} (s,\pi(s)) \right] - (1-\upsilon)\E_{s\sim\rho_0^{\beta_1}} \left[ Q_{\omega}^{\pi} (s,\pi(s)) \right] -\upsilon \E_{s\sim\rho_0^{\beta_2}} \left[ Q_{\omega}^{\pi} (s,\pi(s)) \right] \right\|_1 \\
& \leq (1-\upsilon) \left\| \E_{s\sim\rho_0^{\pi}} \left[ Q_{\omega}^{\pi} (s,\pi(s)) \right] - \E_{s\sim\rho_0^{\beta_1}} \left[ Q_{\omega}^{\pi} (s,\pi(s)) \right] \right\|_1 + \upsilon \left\| \E_{s\sim\rho_0^{\pi}} \left[ Q_{\omega}^{\pi} (s,\pi(s)) \right] -  \E_{s\sim\rho_0^{\beta_2}} \left[ Q_{\omega}^{\pi} (s,\pi(s)) \right] \right\|_1 \\
& \leq 2 \zeta_{\pi} \left( (1-\upsilon) D_{TV}(\rho_0^{\pi},\rho_0^{\beta_1}) + \upsilon D_{TV}(\rho_0^{\pi},\rho_0^{\beta_2}) \right).
\end{split}
\end{equation*}
Since policy $\beta_1$ is embodied by all environment samples stored in the full ERB, $D_{TV}(\rho_0^{\pi},\rho_0^{\beta_1})\approx 0$,
$$
\left\| J(\pi) - \tilde{J}(\pi,\beta_1,\beta_2,\upsilon) \right\|_1 \leq 2 \zeta_{\pi} \upsilon D_{TV}(\rho_0^{\pi},\rho_0^{\beta_2}).
$$

Further note that
$$
\tilde{J}(\pi,\beta_1,\beta_2,\upsilon) - \tilde{J}(\pi,\pi'\beta_1,\beta_2,\upsilon) = \upsilon \E_{\tau\sim\beta_2} \left[ A_{\omega}^{\beta_2} (s_t,\pi(s_t)) - A_{\omega}^{\beta_2} (s_t,\pi'(s_t)) \right].
$$
In view of this,
\begin{equation*}
\begin{split}
& \ \ \ \left\| J(\pi) - \tilde{J}(\pi,\pi',\beta_1,\beta_2,\upsilon) \right\|_1 \\
& \leq \left\| J(\pi) - \tilde{J}(\pi, \beta_1,\beta_2,\upsilon) \right\|_1 + \upsilon \left| \E_{\tau\sim\beta_2} \left[ A_{\omega}^{\beta_2} (s_t,\pi(s_t)) - A_{\omega}^{\beta_2} (s_t,\pi'(s_t)) \right] \right| \\
& \leq 2 \zeta_{\pi} \upsilon D_{TV}(\rho_0^{\pi},\rho_0^{\beta_2}) + \upsilon \left| \E_{\tau\sim\beta_2} \left[
\sum_{t=0}^{\infty} \gamma^t \left( Q_{\omega}^{\pi}(s_t,\pi(s_t)) - Q_{\omega}^{\pi} (s_t,\pi'(s_t)) \right) \right] \right| \\
& \leq 2 \zeta_{\pi} \upsilon D_{TV}(\rho_0^{\pi},\rho_0^{\beta_2}) + \upsilon \E_{\tau\sim\beta_2} \left[ \sum_{t=0}^{\infty} \gamma^t \left| Q_{\omega}^{\pi}(s_t,\pi(s_t)) - Q_{\omega}^{\pi} (s_t,\pi'(s_t)) \right| \right] \\
& \leq 2 \zeta_{\pi} \upsilon D_{TV}(\rho_0^{\pi},\rho_0^{\beta_2}) + \upsilon \E_{\tau\sim\beta_2} \left[ \sum_{t=0}^{\infty} \gamma^t \chi_1 \| \pi(s_t)-\pi'(s_t) \|_1 \right] \\
& \leq 2 \zeta_{\pi} \upsilon D_{TV}(\rho_0^{\pi},\rho_0^{\beta_2}) + \upsilon \E_{\tau\sim\beta_2} \left[ \sum_{t=0}^{\infty} \gamma^t \chi_1\chi_2\alpha  \right] \\
& = 2 \zeta_{\pi} \upsilon D_{TV}(\rho_0^{\pi},\rho_0^{\beta_2}) + \frac{\upsilon\chi_1\chi_2\alpha}{1-\gamma}.
\end{split}
\end{equation*}
This proves Proposition \ref{po-perf-bound}. \QEDB

It is interesting to note that Proposition \ref{po-perf-bound} produces, in theory, a DRL algorithm with monotonic policy improvement guarantees. Specifically, define
\begin{equation*}
\begin{split}
M(\pi,\beta) & =\tilde{J}(\pi,\beta)-2 \zeta_{\pi} D_{TV}(\rho_0^{\pi},\rho_0^{\beta}) \\
& = \E_{s\sim\rho_0^{\beta}} \left[ Q_{\omega}^{\pi}(s,\pi(s)) \right] - 2 \zeta_{\pi} D_{TV}(\rho_0^{\pi},\rho_0^{\beta}).
\end{split}
\end{equation*}
A new policy updating rule can be established. Particularly, given policy $\pi_i$ at the $i$-th learning iteration ($i\geq 0$), policy $\pi_{i+1}$ for the next learning iteration can be determined as
\begin{equation}
\pi_{i+1}=\argmax_{\beta} M(\pi_i,\beta)
\label{equ-lr-theory}
\end{equation}
Based on Proposition \ref{po-perf-bound}, it is straightforward to see that:
$$
J(\pi_{i+1})\geq M(\pi_i,\pi_{i+1}) \geq M(\pi_i,\pi_i) = J(\pi_i).
$$
(\ref{equ-lr-theory}) hence guarantees monotonic policy improvement. Consequently, we can prove the convergence of the policy training process upon using (\ref{equ-lr-theory}) \cite{schulman2015icml}. Since (\ref{equ-lr-theory}) is not practically feasible, we choose to continuously update the elite ERB with elite trajectories newly generated by trained policies. Because the elite ERB embodies policy $\beta_2$ in $\tilde{J}(\pi,\beta_2)$, the updating process can be considered as approximating (\ref{equ-lr-theory}) in an efficient but imprecise manner. In fact, the elite ERB keeps track of those ``\emph{good}" initial states on which the train policies have achieved the highest cumulative rewards. Accordingly, the probability density of $\rho_0^{\beta_2}$ is expected to concentrate mainly on such ``good" states. This will approximately maximize $\E_{s\sim\rho_0^{\beta_2}}\left[ Q_{\omega}^{\pi}(s,\pi(s)) \right]$ and hence $M(\pi,\beta_2)$. Meanwhile, with the help of the VAE generative model $\mathcal{V}$ and the regularizer in (\ref{eq-ev-dgp}), we can effectively train $\pi_{i+1}$ to approximate $\beta_2$, thereby roughly fulfilling the optimization requirement in (\ref{equ-lr-theory}). Following the same idea, we can also construct the following learning rules with guaranteed monotonic policy improvement:
$$
\pi_{i+1}=\argmax_{\beta}\tilde{J}(\pi_i,\beta_1,\beta,\upsilon)-2 \zeta_{\pi_i} \upsilon D_{TV}(\rho_0^{\pi_i},\rho_0^{\beta}),
$$
$$
\pi_{i+1}=\argmax_{\beta} \left( \max_{\pi'}\tilde{J}(\pi_i,\pi',\beta_1,\beta,\upsilon)-2 \zeta_{\pi_i} \upsilon D_{TV}(\rho_0^{\pi_i},\rho_0^{\beta})-\upsilon \left| \E_{\tau\sim\beta}\left[ Q_{\omega}^{\pi}(s_t,\pi(s_t))-Q_{\omega}^{\pi}(s_t,\pi'(s_t))\right] \right| \right),
$$
which can be approximately realized by interpolation merging and two-step merging respectively, providing theoretical evidence regarding why $\nabla^{IM}J(\pi)$ and $\nabla^{2M}J(\pi)$ can be utilized to train policy $\pi$ effectively.

\section*{Appendix C}

This appendix studies the learning bias introduced by imprecise approximation of the value function. Specifically, we distinguish $Q_{\omega}^{\pi}(s,a)$ and $\tilde{Q}_{\omega}^{\pi}(s,a)$ which represent respectively the precise and imprecise approximations of the Q-function $Q^{\pi}(s,a)$ of policy $\pi$ with respect to any state $s\in\mathbb{S}$ and any action $a\in\mathbb{A}$. Define
\begin{equation*}
\begin{split}
& \hat{J}(\pi,\beta_2)=J(\beta_2)+\E_{\tau\sim\beta_2}\left[ \tilde{A}_{\omega}^{\beta_2}(s_t,\pi(s_t)) \right], \\
& \hat{J}(\pi,\beta_1,\beta_2,\upsilon)=(1-\upsilon)\hat{J}(\pi,\beta_1)+\upsilon \hat{J}(\pi,\beta_2), \\
& \hat{J}(\pi,\pi',\beta_1,\beta_2,\upsilon)=(1-\upsilon)\hat{J}(\pi,\beta_1)+\upsilon \hat{J}(\pi',\beta_2),
\end{split}
\end{equation*}
where $\tilde{A}_{\omega}^{\beta}=\tilde{Q}_{\omega}^{\pi}(s,\pi(s))-\tilde{Q}_{\omega}^{\pi}(s,\beta(s))$. Similar to Proposition \ref{po-perf-bound}, their differences from $J(\pi)$ can be bounded according to Proposition \ref{prop-perfv-bound} below.
\begin{proposition}
Assuming the existence of positive constants $\chi_3$ and $\chi_4$, such that
$$
\chi_3=\max_{s\in\mathbb{S},a\in\mathbb{A}} \left| Q_{\omega}^{\pi}(s,a) - \tilde{Q}_{\omega}^{\pi}(s,a)\right| < \infty,
$$
$$
\max_{s\in\mathbb{S}} \left| \tilde{Q}_{\omega}^{\pi}(s,a) - \tilde{Q}_{\omega}^{\pi}(s,a')\right| \leq \chi_4 \|a-a_1\|_1,
$$
whenever actions $a$ and $a'$ are sufficiently close, then the differences of $\hat{J}(\pi,\beta_2)$, $\hat{J}(\pi,\beta_1,\beta_2,\upsilon)$ and $\hat{J}(\pi,\pi',\beta_1,\beta_2,\upsilon)$ from $J(\pi)$ can be bounded as below:
\begin{equation*}
\begin{split}
& \left\| J(\pi)-\hat{J}(\pi,\beta_2) \right\|_1 \leq \zeta_{\pi} D_{TV} (\rho_0^{\pi}, \rho_0^{\beta_2} ) + \frac{2\chi_3}{1-\gamma}, \\
& \left\| J(\pi)- \hat{J}(\pi,\beta_1,\beta_2,\upsilon) \right\|_1 \leq \zeta_{\pi} \upsilon D_{TV}(\rho_0^{\pi},\rho_0^{\beta_2}) + \frac{2\chi_3}{1-\gamma}, \\
& \left\| J(\pi)- \hat{J}(\pi,\pi',\beta_1,\beta_2,\upsilon) \right\|_1 \leq \zeta_{\pi} \upsilon D_{TV}(\rho_0^{\pi},\rho_0^{\beta_2}) + \frac{2\chi_3}{1-\gamma} + \frac{\upsilon\chi_2\chi_4\alpha}{1-\gamma}
\end{split}
\end{equation*}
\label{prop-perfv-bound}
\end{proposition}

{\bf Proof}: Using Proposition \ref{po-perf-bound},
\begin{equation*}
\begin{split}
& \ \ \ \left\| J(\pi)-\hat{J}(\pi,\beta_2) \right\|_1 \\
& \leq \left\| J(\pi)-\tilde{J}(\pi,\beta_2) \right\|_1 + \left\| \E_{\tau\sim\beta_2} \left[ \tilde{A}_{\omega}^{\beta_2}(s,\pi(s)) - A_{\omega}^{\beta_2}(s,\pi(s))\right] \right\|_1 \\
& \leq 2 \zeta_{\pi} D_{TV} (\rho_0^{\pi}, \rho_0^{\beta_2} ) + \left\| \E_{\tau\sim\beta_2} \left[ \tilde{Q}_{\omega}^{\pi}(s,\pi(s)) - Q_{\omega}^{\pi}(s,\pi(s)) \right] \right\|_1 + \left\| \E_{\tau\sim\beta_2} \left[ Q_{\omega}^{\pi}(s,\beta_2(s)) - \tilde{Q}_{\omega}^{\pi}(s,\beta_2(s)) \right] \right\|_1 \\
& \leq 2 \zeta_{\pi} D_{TV} (\rho_0^{\pi}, \rho_0^{\beta_2} ) + \frac{2\chi_3}{1-\gamma}
\end{split}
\end{equation*}
Building on the above, it is straightforward to prove the bounds with respect to $\hat{J}(\pi,\beta_1,\beta_2,\upsilon)$ and $\hat{J}(\pi,\pi',\beta_1,\beta_2,\upsilon)$. The details will be omitted here. \QEDB

Proposition \ref{prop-perfv-bound} shows that the error involved in approximating the Q-function by $\tilde{Q}_{\omega}^{\pi}$, as measured by $\chi_3$, introduces additional bias for policy training. Meanwhile, the extra bias caused by using two-step merging, as reflected by the difference between $J(\pi)$ and $\hat{J}(\pi,\pi',\beta_1,\beta_2,\upsilon)$, can be controlled via the learning rate $\alpha$. Besides Proposition \ref{prop-perfv-bound}, we can also bound directly the bias involved in estimating and merging DPGs. Consider
$$
\nabla J(\pi)=\E_{\tau\sim\pi}\left[ \nabla Q_{\omega}^{\pi}(s_t,\pi(s_t)) \right] \ \mathrm{and}
$$
$$
\nabla^e\hat{J}(\pi)=\E_{\tau\sim\beta_2}\left[ \nabla \tilde{Q}_{\omega}^{\pi}(s_t,\pi(s_t)) \right]
$$
that represent respectively the precise as well as imprecise estimations of the DPG (the latter for elite DPG). In order to bound the difference between the two, we need to utilize an important assumption: the state-transition probability of an MDP is \emph{continuous} with respect to its actions, meaning that small differences in actions performed by an RL agent will lead to small differences in state-transition probabilities, as formulated below:
\begin{equation}
\max_{s,s'\in\mathbb{S}} \left| \Pr(s,s',a)-\Pr(s,s',a') \right|\leq \frac{c}{\|\mathbb{S}\|} \left\| a-a' \right\|_1
\label{eq-sm-st}
\end{equation}
where $c>0$ is a constant. Building on this assumption, we can formulate and prove Lemma \ref{lemma-2} first.
\begin{lemma}
Given two policies $\pi$ and $\beta$, let $\rho_t^{\pi}$ and $\rho_t^{\beta}$ denote respectively the probability distributions that govern the probability for an MDP to be in any state $s\in\mathbb{S}$ at time $t$, provided that an RL agent follows either $\pi$ or $\beta$ to interact with its learning environment since time 0. In line with the continuous state-transition condition in (\ref{eq-sm-st}), the total variation distance between $\rho_t^{\pi}$ and $\rho_t^{\beta}$ can be bounded as below:
\begin{equation}
D_{TV}(\rho_t^{\pi},\rho_t^{\beta})\leq \frac{t c}{2}\Delta_{\pi,\beta}+D_{TV}(\rho_0^{\pi},\rho_0^{\beta}),
\label{equ-rho-bound}
\end{equation}
where $\Delta_{\pi,\beta}=\max_{s\in\mathbb{S}}\| \pi(s)-\beta(s) \|_1$.
\label{lemma-2}
\end{lemma}
\noindent
{\bf Proof}: Consider the state-transition at time $t-1$ by following policies $\pi$ and $\beta$ respectively. Based on $\rho_{t-1}^{\pi}$ and $\rho_{t-1}^{\beta}$, $\rho_t^{\pi}$ and $\rho_t^{\beta}$ can be determined according to:
$$
\rho_t^{\pi}(s')=\sum_{s\in\mathbb{S}} \rho_{t-1}^{\pi}(s)\Pr(s,s',\pi(s)),
\rho_t^{\beta}(s')=\sum_{s\in\mathbb{S}} \rho_{t-1}^{\beta}(s)\Pr(s,s',\beta(s)).
$$
Hence,
\begin{equation*}
\begin{split}
D_{TV}(\rho_t^{\pi},\rho_t^{\beta}) & = \frac{1}{2}\sum_{s'\in\mathbb{S}} \left| \sum_{s\in\mathbb{S}} \rho_{t-1}^{\pi}(s)\Pr(s,s',\pi(s)) - \sum_{s\in\mathbb{S}} \rho_{t-1}^{\beta}(s)\Pr(s,s',\beta(s)) \right| \\
& = \frac{1}{2}\sum_{s'\in\mathbb{S}} \left|
\begin{array}{l}
\sum_{s\in\mathbb{S}} \rho_{t-1}^{\pi}(s)\Pr(s,s',\pi(s)) - \sum_{s\in\mathbb{S}} \rho_{t-1}^{\pi}(s)\Pr(s,s',\beta(s)) + \\ \sum_{s\in\mathbb{S}} \rho_{t-1}^{\pi}(s)\Pr(s,s',\beta(s)) - \sum_{s\in\mathbb{S}} \rho_{t-1}^{\beta}(s)\Pr(s,s',\beta(s))
\end{array}
\right|\\
& \leq \frac{1}{2} \sum_{s'\in\mathbb{S}} \left(
\left| \sum_{s\in\mathbb{S}} \rho_{t-1}^{\pi}(s)(\Pr(s,s',\pi(s)) - \Pr(s,s',\beta(s))) \right| \right. \\
& \left. \ \ \ +  \left| \sum_{s\in\mathbb{S}} (\rho_{t-1}^{\pi}(s) - \rho_{t-1}^{\beta}(s)) \Pr(s,s',\beta(s))
\right| \right) \\
& \leq \frac{1}{2} \sum_{s'\in\mathbb{S}} \left(
\sum_{s\in\mathbb{S}} \rho_{t-1}^{\pi}(s) \left| (\Pr(s,s',\pi(s)) - \Pr(s,s',\beta(s))) \right| \right. \\
& \left. \ \ \ + 
\sum_{s\in\mathbb{S}} \left|(\rho_{t-1}^{\pi}(s) - \rho_{t-1}^{\beta}(s)) \right| \Pr(s,s',\beta(s))
\right) \\
& = \frac{1}{2} \sum_{s\in\mathbb{S}} \rho_{t-1}^{\pi}(s) \sum_{s'\in\mathbb{S}} \left| (\Pr(s,s',\pi(s)) - \Pr(s,s',\beta(s))) \right| \\
& \ \ \ + \frac{1}{2} \sum_{s\in\mathbb{S}} \left|(\rho_{t-1}^{\pi}(s) - \rho_{t-1}^{\beta}(s)) \right| \sum_{s'\in\mathbb{S}} \Pr(s,s',\beta(s)) \\
& \leq \frac{1}{2} \sum_{s\in\mathbb{S}} \rho_{t-1}^{\pi}(s)  \sum_{s'\in\mathbb{S}} \frac{c}{\|\mathbb{S}\|} \Delta_{\pi,\beta} + D_{TV}(\rho_{t-1}^{\pi}, \rho_{t-1}^{\beta}) \\
& = \frac{c}{2} \Delta_{\pi,\beta} + D_{TV}(\rho_{t-1}^{\pi}, \rho_{t-1}^{\beta})
\end{split}
\end{equation*}
In view of the above, it can be shown further that
\begin{equation*}
\begin{split}
D_{TV}(\rho_t^{\pi},\rho_t^{\beta}) & \leq \frac{c}{2} \Delta_{\pi,\beta} + D_{TV}(\rho_{t-1}^{\pi}, \rho_{t-1}^{\beta}) \\
& \leq \frac{c}{2} \Delta_{\pi,\beta} + \left( \frac{c}{2} \Delta_{\pi,\beta} + D_{TV}(\rho_{t-2}^{\pi}, \rho_{t-2}^{\beta}) \right)\\
& \leq \ldots \\
& \leq \frac{t c}{2} \Delta_{\pi,\beta} + D_{TV}(\rho_0^{\pi}, \rho_0^{\beta})
\end{split}
\end{equation*}
This proves Lemma \ref{lemma-2}. \QEDB

By leveraging on Lemma \ref{lemma-2}, Proposition \ref{prop-perf-v-bound} can be established to bound the difference between  $\nabla J(\pi)$ and $\nabla^e \hat{J}(\pi)$.
\begin{proposition}
Define
$$
\psi_1=\max_{s\in\mathbb{S}}\left\| \nabla Q_{\omega}^{\pi} (s_t,\pi(s_t)) - \nabla \tilde{Q}_{\omega}^{\pi} (s_t,\pi(s_t)) \right\|_1<\infty \ \mathrm{and}
$$
$$
\psi_2=\max_{s\in\mathbb{S}} \left\| \nabla \tilde{Q}_{\omega}^{\pi} (s_t,\pi(s_t)) \right\|_1 < \infty.
$$
Based on the continuous state-transition condition in (\ref{eq-sm-st}), the difference between $\nabla J(\pi)$ and $\nabla \hat{J}(\pi)$ is bounded as below:
\begin{equation}
\|\nabla J(\pi) - \nabla^e \hat{J}(\pi)\|_1 \leq \frac{\psi_1}{1-\gamma} + \frac{\gamma c\psi_2}{(1-\gamma)^2}\Delta_{\pi,\beta_2}+ \frac{2\psi_2}{1-\gamma} D_{TV}(\rho_0^{\pi},\rho_0^{\beta_2}).
\label{equ-bound-edpg}
\end{equation}
\label{prop-perf-v-bound}
\end{proposition}

{\bf Proof}:
\begin{equation*}
\begin{split}
& \ \ \ \|\nabla J(\pi) - \nabla^e \hat{J}(\pi)\|_1 \\
& = \left\| \E_{\tau\sim\pi} \left[ \nabla Q_{\omega}^{\pi} (s_t,\pi(s_t)) - \nabla \tilde{Q}_{\omega}^{\pi} (s_t,\pi(s_t)) \right] + \E_{\tau\sim\pi} \left[ \nabla \tilde{Q}_{\omega}^{\pi} (s_t,\pi(s_t))\right] - \E_{\tau\sim\beta_2} \left[ \nabla \tilde{Q}_{\omega}^{\pi} (s_t,\pi(s_t)) \right] \right\|_1 \\
& \leq \left\| \sum_{t=0}^{\infty} \gamma^t \sum_{s\in\mathbb{S}} \rho_t^{\pi}(s) \left( \nabla Q_{\omega}^{\pi} (s_t,\pi(s_t)) - \nabla \tilde{Q}_{\omega}^{\pi} (s_t,\pi(s_t)) \right) \right\|_1 + \left\| \sum_{t=0}^{\infty} \gamma^t \sum_{s\in\mathbb{S}}\left( \rho_t^{\pi}(s)-\rho_t^{\beta_2}(s) \right) \tilde{Q}_{\omega}^{\pi} (s_t,\pi(s_t)) \right\|_1 \\
& \leq \psi_1 \sum_{t=0}^{\infty} \gamma^t + 2 \psi_2 \sum_{t=0}^{\infty} \gamma^t D_{TV} (\rho_t^{\pi}, \rho_t^{\beta_2}) \\
& \leq \frac{\psi_1}{1-\gamma} + 2 \psi_2 \sum_{t=0}^{\infty} \gamma^t \frac{t c}{2}\Delta_{\pi,\beta_2} + 2\psi_2 \sum_{t=0}^{\infty}\gamma^t D_{TV} (\rho_0^{\pi},\rho_0^{\beta_2}) \\
& = \frac{\psi_1}{1-\gamma} + \frac{\gamma c\psi_2}{(1-\gamma)^2}\Delta_{\pi,\beta_2}+ \frac{2\psi_2}{1-\gamma} D_{TV}(\rho_0^{\pi},\rho_0^{\beta_2}).
\end{split}
\end{equation*}
This proves Proposition \ref{prop-perf-v-bound}. \QEDB

Proposition \ref{prop-perf-v-bound} indicates that the bias introduced by $\nabla^e \hat{J}(\pi)$ can be mitigated by reducing either $\Delta_{\pi,\beta_2}$ or $D_{TV}(\rho_0^{\pi},\rho_0^{\beta_2})$. This is consistent with the findings captured by Proposition \ref{po-perf-bound}. Particularly, Proposition \ref{prop-perf-v-bound} provides theoretical clues regarding the importance of adopting the regularizer in (\ref{eq-e-dgp}) and (\ref{eq-ev-dgp}) that helps to keep $\Delta_{\pi,\beta_2}$ at a low level. As a special case of Proposition \ref{prop-perf-v-bound} when $Q_{\omega}^{\pi}=\tilde{Q}_{\omega}^{\pi}$ and hence $\psi_1\approx 0$,
$$
\|\nabla J(\pi) - \nabla^e \hat{J}(\pi)\|_1 \leq \frac{\gamma c\psi_2}{(1-\gamma)^2}\Delta_{\pi,\beta_2}+ \frac{2\psi_2}{1-\gamma} D_{TV}(\rho_0^{\pi},\rho_0^{\beta_2}),
$$
which again confirms the importance of controlling $\Delta_{\pi,\beta_2}$ via the regularizer in (\ref{eq-e-dgp}) and (\ref{eq-ev-dgp}), even when the Q-function can be approximated precisely. In line with the above analysis, define
$$
\nabla^{IM} \hat{J}(\pi)=(1-\upsilon)\E_{\tau\sim\beta_1}\left[ \nabla\tilde{Q}_{\omega}^{\pi} (s_t,\pi(s_t)) \right] + \upsilon \E_{\tau\sim\beta_2}\left[ \nabla\tilde{Q}_{\omega}^{\pi}(s_t,\pi(s_t)) \right],
$$
which corresponds to the DPG obtained via interpolation merging. Its difference from $\nabla J(\pi)$ can be bounded below:
\begin{equation}
\left\| \nabla J{\pi} - \nabla^{IM} \hat{J}(\pi) \right\|_1 \leq
\frac{\psi_1}{1-\gamma} + \frac{(1-\upsilon)\gamma c\psi_2}{(1-\gamma)^2}\Delta_{\pi,\beta_1} + \frac{\upsilon\gamma c\psi_2}{(1-\gamma)^2}\Delta_{\pi,\beta_2} + \frac{2\upsilon\psi_2}{1-\gamma} D_{TV} (\rho_0^{\pi},\rho_0^{\beta_2}).
\label{equ-bound-imdpg}
\end{equation}
In comparison to (\ref{equ-bound-edpg}), the bound on the learning bias in (\ref{equ-bound-imdpg}) can be adjusted through $\upsilon$. Since policy $\beta_1$ is expected to be more similar to policy $\pi$ than $\beta_2$, i.e., $\Delta_{\pi,\beta_1}\leq\Delta_{\pi,\beta_2}$, with sufficiently small $\upsilon$,
$$
\left\| \nabla J(\pi) - \nabla^{IM} \hat{J}(\pi) \right\|_1 \leq \left\| \nabla J(\pi) - \nabla^e \hat{J}(\pi) \right\|_1,
$$
confirming again that $\nabla^{IM} J(\pi)$ can mitigate the performance impact of using elite DPG for policy training. Further define
$$
\nabla^{2M} \hat{J}(\pi)=(1-\upsilon)\E_{\tau\sim\beta_1}\left[ \nabla\tilde{Q}_{\omega}^{\pi} (s_t,\pi(s_t)) \right] + \upsilon \E_{\tau\sim\beta_2}\left[ \nabla\tilde{Q}_{\omega}^{\pi}(s_t,\pi'(s_t)) \right],
$$
which corresponds to the DPG obtained via two-step merging. Its difference from $\nabla J(\pi)$ can be bounded below:
\begin{equation}
\left\| \nabla J(\pi) - \nabla^{2M} \hat{J}(\pi) \right\|_1 \leq \left\| \nabla J(\pi) - \nabla^{IM} \hat{J}(\pi) \right\|_1
+ \frac{\psi_3\chi_2\alpha}{1-\gamma},
\label{equ-bound-2mdpg}
\end{equation}
under the assumption that, when $\alpha$ is sufficiently small, there exists a positive constant $\psi_3$ such that:
$$
\left\| \nabla \tilde{Q}_{\omega}^{\pi} (s,\pi(s)) - \nabla \tilde{Q}_{\omega}^{\pi} (s,\pi'(s)) \right\|_1 < \psi_3 \left\| \pi(s)-\pi'(s) \right\|_1.
$$
(\ref{equ-bound-2mdpg}) suggests that $\nabla^{2M} J(\pi)$ can introduce extra learning bias than $\nabla^{IM} J(\pi)$ for policy training. Nevertheless, the additional bias can be controlled via $\alpha$. With small enough $\alpha$,
$$
\left\| \nabla J(\pi) - \nabla^{2M} \hat{J}(\pi) \right\|_1 \leq \left\| \nabla J(\pi) - \nabla^e \hat{J}(\pi) \right\|_1.
$$
Hence, similar to $\nabla^{IM} J(\pi)$, $\nabla^{2M} J(\pi)$ can also mitigate the performance impact of using elite DPGs for policy training. Moreover, noisy quadratic analysis in Proposition \ref{po-nqa} suggests that $\nabla^{2M} J(\pi)$ can effectively reduce the variance of trained policy parameters.

\section*{Appendix D}

This appendix presents a proof of Proposition \ref{po-nqa}.

{\bf Proof}: Start with the learning rule for updating $\theta^c$ in (\ref{eq-lr}). Considering the $i$-th learning iteration with arbitrary $i\geq 1$ and taking the expectation at both sides of the learning rule, we obtain
$$
\E\left[ \theta^c_{i+1} \right] = (I-\alpha A) \E\left[ \theta^c_i \right].
$$
\noindent
Given the assumption that $\alpha$ is small so that $Diag(\alpha A)<Diag(I)$, the equation above is a \emph{contraction mapping} from $\E\left[ \theta^c_i \right]$ to $\E\left[ \theta^c_{i+1} \right]$. Hence $\E\left[ \theta^c_i \right]$ converges to $\theta^*=0$ as $i$ approaches to $\infty$. Therefore $\theta^c$ is trained bias-free. Furthermore,
\begin{equation*}
\begin{split}
\V\left[ \theta^c_{i+1} \right] & = \E\left[ \left( (I-\alpha A)(\theta^c_i-\E[\theta^c_i]) + \alpha A c_1 \right)^2  \right] \\
& = (I-\alpha A)^2 \V\left[ \theta^c_i \right] + \alpha^2 A^2 \Sigma_1.
\end{split}
\end{equation*}
\noindent
With $i$ approaching to $\infty$, the fixed point of $\V\left[ \theta^c_{\infty} \right]$ must satisfy
$$
\V\left[ \theta^c_{\infty} \right]=(I-\alpha A)^2 \V\left[ \theta^c_{\infty} \right] + \alpha^2 A^2 \Sigma_1.
$$
\noindent
Solving the above equation gives rise to
$$
\V\left[ \theta^c_{\infty} \right]=\alpha^2 \left[ I-(I-\alpha A)^2 \right]^{-1} A^2 \Sigma_1.
$$

Following the same procedure, we can proceed to analyze the learning rule for $\theta^{IM}$ in (\ref{eq-lr}). Specifically,
$$
\E\left[ \theta^{IM}_{i+1} \right]=(I-\alpha A) \E\left[ \theta^{IM}_i \right]+\alpha\upsilon A\epsilon.
$$
\noindent
We can re-write the above as
$$
\E\left[ \theta^{IM}_{i+1} \right]-\upsilon\epsilon=(I-\alpha A) \left( \E\left[ \theta^{IM}_i \right]-\upsilon\epsilon \right).
$$
\noindent
This is a contraction mapping from $\left( \E\left[ \theta^{IM}_i \right]-\upsilon\epsilon \right)$ to $\left(\E\left[ \theta^{IM}_{i} \right]-\upsilon\epsilon\right)$ that converges to 0. As a result, when $i\rightarrow\infty$, $\E\left[ \theta^{IM}_{i} \right]$ approaches to $\upsilon\epsilon$. Meanwhile, we can determine $\V\left[ \theta^{IM}_{\infty} \right]$ as follows:
\begin{equation*}
\begin{split}
\V\left[ \theta^{IM}_{i+1} \right] & = \E\left[ \left( (I-\alpha A)(\theta^{IM}_i-\E[\theta^{IM}_i]) + \alpha (1-\upsilon) A c_1 + \alpha\upsilon A (c_2-\epsilon) \right)^2  \right] \\
& = (I-\alpha A)^2 \V\left[ \theta^{IM}_i \right] + \alpha^2 (1-\upsilon)^2 A^2 \Sigma_1 + \alpha^2\upsilon^2 A^2\Sigma_2.
\end{split}
\end{equation*}
\noindent
By solving this equation, we obtain
$$
\V[\theta_{\infty}^{IM}] = \alpha^2 \left[ I-(I-\alpha A)^2 \right]^{-1}  A^2 \left( (1-\upsilon)^2 \Sigma_1 + \upsilon^2 \Sigma_2 \right).
$$

Finally the analysis on the learning rule for $\theta^{2M}$ in (\ref{eq-lr}) can be conducted in the same way. Particularly,
$$
\E\left[ \theta^{2M}_{i+1} \right]=(I-\alpha\upsilon A) (I-\alpha(1-\upsilon) A) \E\left[ \theta^{2M}_i \right]+\alpha \upsilon A\epsilon.
$$
\noindent
Let $\Omega=\left[ I-(I-\alpha\upsilon A) (I-\alpha(1-\upsilon)A) \right]^{-1} \alpha\upsilon$, the above can be re-written as
\begin{equation*}
\begin{split}
\E\left[ \theta^{2M}_{i+1} \right] - \Omega A \epsilon & = (I-\alpha\upsilon A) (I-\alpha(1-\upsilon)A) \E\left[ \theta^{2M}_i \right]+\alpha \upsilon A\epsilon - \Omega A \epsilon \\
& = (I-\alpha\upsilon A) (I-\alpha(1-\upsilon)A) \left( \E\left[ \theta^{2M}_i \right] - \Omega A \epsilon \right).
\end{split}
\end{equation*}
\noindent
Clearly, $Diag((I-\alpha\upsilon A) (I-\alpha(1-\upsilon)A))<Diag(I)$ and the mapping above is hence a contraction. This implies that $\E\left[ \theta^{2M}_i \right]$ converges to $\Omega A \epsilon$ with $i\rightarrow\infty$. In the meantime, the fixed point of $\V\left[ \theta^{2M} \right]$ can be determined by solving the equation below:
\begin{equation*}
\begin{split}
\V\left[ \theta^{2M}_{i+1} \right] & = \E\left[ \left(
\begin{array}{l}
(I-\alpha\upsilon A) (I-\alpha(1-\upsilon) A) (\theta^{2M}_i-\E[\theta^{2M}_i]) + \alpha (1-\upsilon)A (I-\alpha\upsilon A) c_1 \\
+ \alpha \upsilon A (c_2-\epsilon)
\end{array}
\right)^2  \right] \\
& = (I-\alpha\upsilon A)^2 (I-\alpha(1-\upsilon) A)^2 \V\left[ \theta^{IM}_i \right] + \alpha^2 (1-\upsilon)^2 A^2 (I-\alpha\upsilon A)^2 \Sigma_1 + \alpha^2 \upsilon^2 A^2 \Sigma_2.
\end{split}
\end{equation*}
\noindent
Consequently, we have
$$
\V[\theta_{\infty}^{2M}]= \alpha^2 \left[ I- (I-\alpha\upsilon A)^2 (I-\alpha(1-\upsilon)A)^2 \right]^{-1} A^2 \left( (1-\upsilon)^2 (I-\alpha\upsilon A)^2\Sigma_1 + \upsilon^2 \Sigma_2 \right).
$$
This proves Proposition \ref{po-nqa}. \QEDB

\section*{Appendix E}

This appendix provides detailed information about hyper-parameter settings of four competing algorithms, i.e., SAC, PPO, TD3 and IPG, which have been evaluated empirically in Section \ref{sec-ex}. The hyper-parameter settings of TD3-IM and TD3-2M are identical to those of TD3. For all algorithms, the value function network and the policy network have been implemented as DNNs with two hidden layers. A total of 128 ReLU hidden units have been deployed into each hidden layer. This is a commonly used network architecture that has been exploited to solve many challenging continuous action benchmarks \cite{haarnoja2018icml}. The Adam optimizer was also used consistently to train both the value function network and the policy network in all competing algorithms \cite{kingma2014}.

For SAC, the reward discount factor $\gamma=0.99$. The maximum size of the replay buffer is 1M samples. We checked several different learning rate settings ranging from 0.0001 to 0.003 and found that the algorithm can achieve reliable performance over all benchmarks by setting the learning rate to 0.001. We also set the size of the batch of samples used for training the actor and the critic, i.e., $\|\mathcal{B}\|$, to 100. In fact SAC is not sensitive to this hyper-parameter and exhibited similar performance with other optional batch sizes including 64, 128 and 256. On the other hand, as demonstrated in \cite{haarnoja2018icml}, the performance of SAC is heavily influenced by the entropy regularization factor. Hence we tested several different settings of this hyper-parameter, including 0.2, 0.5, 1.0, 2.0, 5.0, and 10.0. The best performance of SAC achieved with respect to the most suitable setting of the entropy regularzation factor on every benchmark has been reported in Section \ref{sec-ex}.

Similar to SAC, in PPO, the reward discount factor $\gamma=0.99$. Each learning iteration is performed on 4000 newly collected environment samples. The learning rate setting for PPO is different from that of SAC. In particular, in order for PPO to learn reliably, the learning rate for training the value function is set to 0.001, which is slightly larger than the learning rate of 0.0003 utilized for training the policy. The clipping factor of PPO is set to either 0.1 or 0.2, subject to the corresponding performance on the respective benchmarks. Only the better performance witnessed in between the two alternative settings has been reported in Section \ref{sec-ex}. Meanwhile, PPO in our experiments adopted the Generalized Advantage Estimation (GAE) technique introduced in \cite{schulman2015} with hyper-parameter $\lambda=0.95$.

As for TD3, following \cite{fujimoto2018}, besides setting $\gamma$ to 0.99 and the maximum replay buffer size to 1M samples, the learning rate for both the actor and the critic has been set to 0.001.  We checked other possible settings of the learning rate, ranging from 0.0001 to 0.003, and found that TD3 can achieve the best tradeoff between learning performance and reliability when the learning rate equals to 0.001. Different from SAC, the batch size in TD3 is 256. However, setting the batch size to 100 as recommended in \cite{fujimoto2018} will not noticeably bring the average learning performance down. Nevertheless, with the batch size of 256, TD3 appears to perform slightly more reliably. In addition to the above, we set the standard deviation for $\mu_a$ in (\ref{eq-sp}) to 0.2. This setting can slightly improve learning reliability in comparison to the recommended setting of 0.1 in \cite{fujimoto2018}. Meanwhile, the standard deviation of the action sampling noise for the target action smoothing mechanism is set to 0.2. The noise is also clipped between -0.5 and 0.5.

Most of the hyper-parameter settings of IPG are similar to those of PPO. Nevertheless, there are some notable differences. Particularly, $\lambda=0.97$ for GAE. This is shown to produce slightly better performance than setting $\lambda$ to 0.95. The batch size for policy and value function update is 64. We tested several different settings of the interpolation coefficient $\nu$, including 0.0, 0.2, 0.5 and 0.8. It seems that IPG can achieve consistently better performance when $\nu=0.2$. Experiment results corresponding to this setting have been reported in Section \ref{sec-ex}.

\end{document}